\documentclass[10pt,journal,compsoc]{IEEEtran}
\ifCLASSOPTIONcompsoc
  \usepackage[nocompress]{cite}
\else
  \usepackage{cite}
\fi
\usepackage{times}
\usepackage{epsfig}
\usepackage{graphicx}
\usepackage{amsmath}
\usepackage{amssymb}
\usepackage{gensymb}
\usepackage{multirow}
\usepackage{hhline}

\usepackage[justification=centering]{caption}

\newcommand{\tabincell}[2]{\begin{tabular}{@{}#1@{}}#2\end{tabular}}

\usepackage{xcolor}
\begin{document}
\title{Glissando-Net: Deep sinGLe vIew category level poSe eStimation ANd 3D recOnstruction}

\author{Bo~Sun,
        Hao~Kang,
        Li~Guan,
        Haoxiang~Li,
        Philippos~Mordohai, Senior Member, IEEE
        and~Gang~Hua, Fellow, IEEE
\IEEEcompsocitemizethanks{
%\IEEEcompsocthanksitem B. Sun was with Stevens Institute of Technology, Hoboken, New Jersey, 07030.
\IEEEcompsocthanksitem B. Sun is with Adobe Inc, San Jose,
California, 95110.\protect\\
E-mail: bosu@adobe.com\\
The work was done when Bo Sun was at Stevens Institute of Technology and Wormpex.
\IEEEcompsocthanksitem P. Mordohai is with Stevens Institute of Technology, Hoboken, New Jersey, 07030.
\IEEEcompsocthanksitem G. Hua is with Dolby Laboratories, Bellevue, Washington, 98004. % <-this % stops an unwanted space
\IEEEcompsocthanksitem H. Kang is with ByteDance Inc., Bellevue, Washington, 98004.
\IEEEcompsocthanksitem H. Li is with Pixocial Technology, Bellevue, Washington, 98004.
\IEEEcompsocthanksitem L. Guan is with Meta Reality Labs, Menlo Park, California, 94025.}}
\IEEEtitleabstractindextext{%
\begin{abstract}
We present a deep learning model, dubbed Glissando-Net, to simultaneously estimate the pose and reconstruct the 3D shape of objects at the category level from a single RGB image. Previous works predominantly focused on either estimating poses (often at the instance level), or reconstructing shapes, but not both. Glissando-Net is composed of two auto-encoders that are jointly trained, one for RGB images and the other for point clouds. We embrace two key design choices in Glissando-Net to achieve a more accurate prediction of the 3D shape and pose of the object given a single RGB image as input. First, we augment the feature maps of the point cloud encoder and decoder with transformed feature maps from the image decoder, enabling effective 2D-3D interaction in both training and prediction. Second, we predict both the 3D shape and pose of the object in the decoder stage. This way, we better utilize the information in the 3D point clouds presented only in the training stage to train the network for more accurate prediction. We jointly train the two encoder-decoders for RGB and point cloud data to learn how to pass latent features to the point cloud decoder during inference. In testing, the encoder of the 3D point cloud is discarded. The design of Glissando-Net is inspired by codeSLAM. Unlike codeSLAM, which targets 3D reconstruction of scenes, we focus on pose estimation and shape reconstruction of objects, and directly predict the object pose and a pose invariant 3D reconstruction without the need of the code optimization step. Extensive experiments, involving both ablation studies and comparison with competing methods,  demonstrate the efficacy of our proposed method, and compare favorably with the state-of-the-art.
\end{abstract}

}

\maketitle

\IEEEdisplaynontitleabstractindextext

% \fi
\IEEEpeerreviewmaketitle

\IEEEraisesectionheading{\section{Introduction}\label{sec:introduction}}

\IEEEPARstart{R}{econstruction} of 3D geometry from images is one of the classical challenging problems in computer vision. 
Recently, researchers were able to recover the depth of the scene~\cite{bloesch2018codeslam}, 3D shape~\cite{fan2017pointsetgeneration}, or object pose~\cite{georgakis2019learning} from a single RGB image. The monocular setting provides only partial observations and makes 3D understanding especially challenging. An effective method must rely on proper prior information, most likely learned from data.

\begin{figure}[bh]
    \centering
    \includegraphics[width=.9\linewidth]{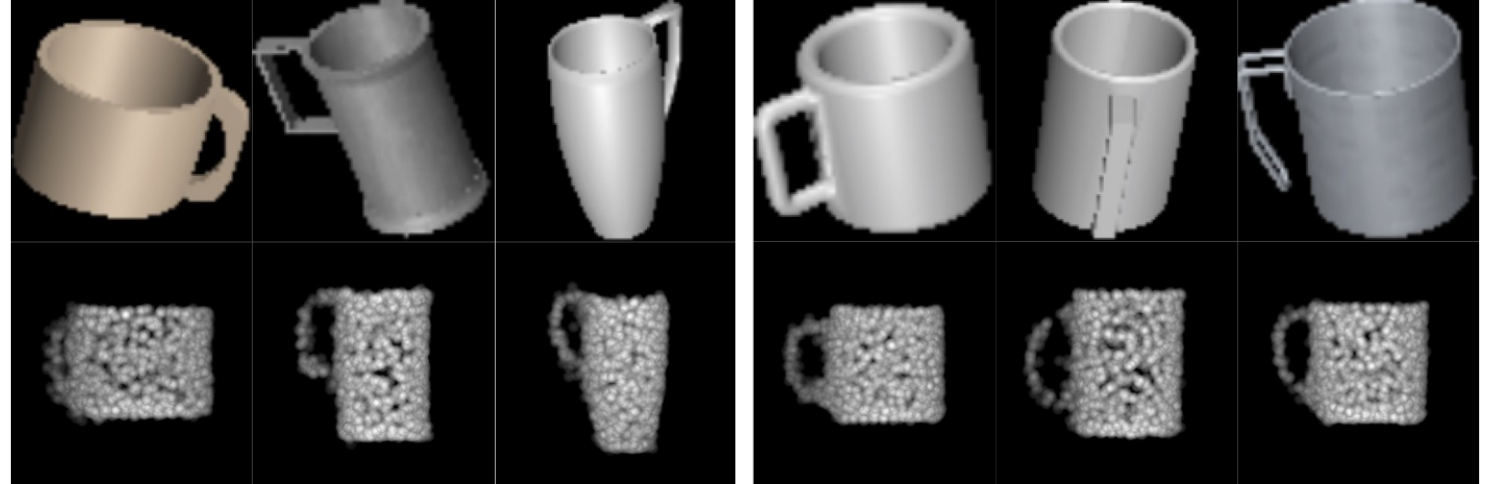}
    \caption{Category-level 3D reconstruction: on the top are input images from NOCS dataset \cite{wang2019normalized} and on the bottom are predicted 3D shapes; on the left are examples of training instances and on the right are testing instances. Our model predicts complete 3D shapes of instances unknown during training based on their RGB appearance and category shape prior learned from data.}
    \label{fig:cup}
\end{figure}

In this paper, from a single RGB image, we jointly estimate the 3D shape and 6D pose of an object. (Pose in this paper always refers to 6D pose.) Unlike existing works that require data from the exact instance when training the model~\cite{kehl2017ssd,xiang2018posecnn}, our method works at the \emph{category level}, observing only objects in the same category, but never the test object in training.
%Trained on objects from the same category, 
%but not the exact object itself, 
It can predict the 3D shape of a novel object and simultaneously estimate its pose. % with respect to the canonical frame.
This setting is under-explored on real data, due to its difficulty, and has broader practical applications. It also poses unique and interesting technical challenges since the model needs to capture both high-level semantic differences across categories and low-level geometric details between instances of the same category. We propose a deep CNN framework to overcome the challenge and name it Glissando-Net, short for deep sinGLe vIew category level poSe eStimation ANd 3D recOnstruction.

In training, our network takes RGB images and paired canonical point clouds to learn the point clouds (3D object shapes) and object poses. The RGB image is processed by a U-Net style RGB encoder-decoder network and the point cloud is processed by a PointNet++ encoder and multiple layers of fully-connected layers as the decoder. We carefully design a feature transform module to facilitate interactions between the RGB features and point cloud features. Our model predicts the canonical 3D point cloud and estimates the pose based on the output features of the point cloud decoder, leveraging the fused 2D and 3D information. 

In testing, we take a single RGB image as input. The point cloud decoder leverages the RGB features concatenated through the feature transform module to generate object shapes and predict their poses. Our model consumes an RGB image with a single centered object. Given an image with multiple objects, we assume the availability of an object detector to localize and crop the objects. 
As shown in Figure~\ref{fig:cup}, without seeing the same instances in training, our model predicts distinguishable 3D shapes and well-aligned object poses. We experimentally compare our method with recent works on two real datasets and observe consistent improvements in both shape reconstruction and pose estimation. 

Our contributions are three-fold: 1) a novel framework for category-level pose estimation and 3D reconstruction from a single RGB image that learns from 3D data during training but does not require them during testing; 2) a feature transform module that enhances 2D-3D interaction during learning;
%that may be applicable beyond the scope of this paper; 
3) better performance than the state-of-the-art in both pose estimation and 3D reconstruction on recent benchmarks.

%------------------------------------------------------------------------
\begin{figure*}[t]
%\vspace{-0.2in}
    \centering
    \includegraphics[width=0.92\linewidth]{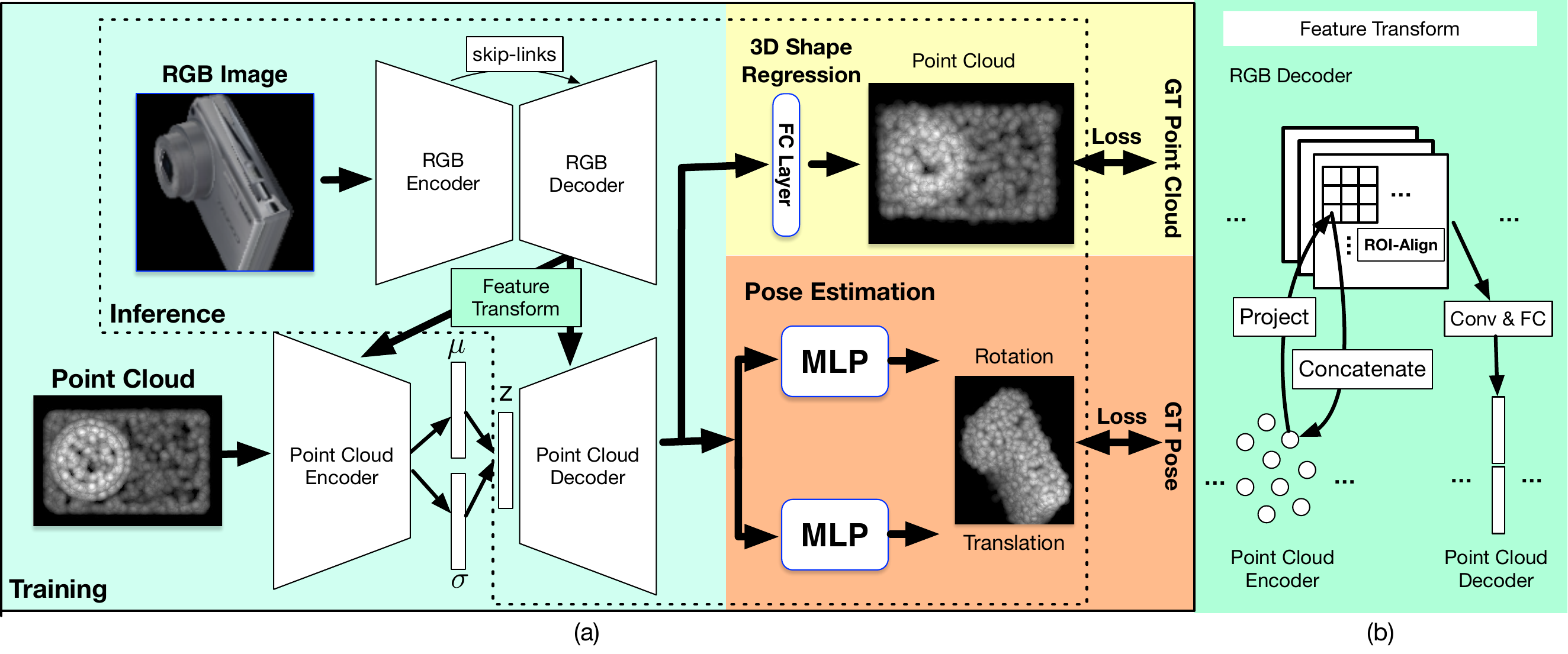}
    %\vspace{-5pt}
    \caption{Overview of Glissando-Net: the RGB image is processed by an encoder-decoder; the feature maps from the RGB decoder are then transformed and concatenated with features in the point cloud decoder, which takes in a  latent code $\mathbf{z}$; the outputs from the point-cloud decoder are processed by a fully-connect layer to regress the 3D shape in canonical pose and two independent multilayer perceptrons (MLP) to estimate its pose; in training, the ground-truth canonical shape is an additional input to a point cloud encoder to learn the distribution of $\mathbf{z}$ and feature maps from the RGB decoder are transformed to augment point cloud features. }
    %\vspace{-10pt}
    \label{fig:workflow}
\end{figure*}

%------------------------------------------------------------------------
\section{Related Work} \label{sec:related}
%PM: check list in backup related.tex to see if anything was missed

Object shape and pose recovery are crucial for many applications, such as 3D reconstruction, robotics, autonomous driving, augmented reality etc. Although this line of work originated from purely geometric methods, learning based approaches recently have shown superior capability to model shape priors. Given the scope of our work, we focus on only discussing the learning-based approaches in this section. What makes the problem challenging, is that the two sub-problems (i.e. the shape recovery problem and object pose recovery problem) are both ill-conditioned. Here, we only discuss learning-based approaches starting with those that are most closely related to ours.

To our knowledge, the only approach that shares all challenges we address, specifically (i) joint shape and pose estimation, (ii) from a single real RGB image, (iii) at the category level, is CPS \cite{manhardt2020cps}. %, which remains unpublished in a referred venue. 
CPS minimizes a loss comprising (i) the symmetric Chamfer distance between the ground truth point cloud and the estimated point cloud transformed according to the estimated pose, and (ii) a regularization term that penalized deviations from the learned class distributions in latent space. This loss measures shape alignment directly without separating shape and pose estimation. 
Recently, the same authors presented CPS++ \cite{manhardt2020cpsv3}, which undergoes fully supervised training on synthetic data, and is then finetuned on real RGB-D data using differentiable rendering techniques for self-supervision.
Unlike our approach, CPS does not extract information from the 3D shapes during training and makes all predictions using the encoder. Moreover, CPS relies on a separate latent space per class, while objects from all classes are encoded in the same latent space by Glissando-Net. 

OLD-Net \cite{fan2022object} takes only RGB images as input for category-level 6D object pose estimation. They propose to directly predict object-level depth from a monocular RGB image by deforming the category-level shape prior into object-level depth and the canonical NOCS representation. Langer et al.\cite{langer2021leveraging} use cross-domain keypoint matches from an RGB image to a rendered CAD model to do object pose predictions and shape estimation. They retrieve multiple candidate CAD models based on a target RGB image at test time and perform matching points between the real image and the retrieved CAD model. Then, these matches are used to calculate the pose and modify the shape of the retrieved CAD model to better fit the object observed. NOCE \cite{lee2021category} proposes a framework that jointly estimates a metric scale shape and pose from a single RGB image. They use both the RGB image and some depth information as input. Glissando-Net, however, does not use any depth information at test time.

A milestone in this line of research was presented by Tulsiani et al. \cite{tulsiani2018} who use multiple views of the object to minimize a geometric loss during training. A single image is sufficient to infer pose and a low-resolution $32\times 32 \times 32$ occupancy grid. 
Insafutdinov and Dosovitskiy \cite{insafutdinov2018unsupervised} also train a network on multiple images, but they employ a point cloud representation and an ensemble of pose predictors to mitigate pose ambiguity. Both \cite{insafutdinov2018unsupervised,tulsiani2018} have been tested only on synthetic data.
In addition to the Pix3D dataset, which we use in our experiments, Sun et al. \cite{sun2018pix3d} present a method for estimating voxelized shape and pose, adopting the 2.5-D sketch as an intermediate representation. 
Nie et al. \cite{nie2020total3dunderstanding} estimate room layout, camera pose, bounding boxes and meshes for the 3D objects from a single image benefiting from the synergies among all tasks. CoReNet \cite{popov2020corenet} addresses the same problem by jointly estimating the class, pose and shape of all objects, instead of each object separately. % PM trained on 80% of instances. No quant on Pix3D
Zhang et al. \cite{Zhang_2021_CVPR_holistic} also address joint inference of scene layout, object poses and meshes via an implicit representation that avoids intersecting surfaces and a graph convolutional network that refines the predictions based on context.

%\PMcomments{[PM: none of the above papers is clear on instance (total3D) vs category-based applicability. Tulsiani, Insaf per-category model, synthetic only.]}
%
GSNet \cite{ke2020gsnet} estimates the 3D pose and  detailed shape of cars from a single image leveraging context, but is limited to a single class of objects.
All of these methods train a separate model for each category, making them susceptible to classification errors by the detector. Glissando-Net relies on a shared latent space across all categories.

Wang et al. \cite{wang2019normalized} introduce the Normalized Object Coordinate Space (NOCS), which is a shared canonical representation for all instances in a category. The network infers the mapping from pixels in the RGB-D input to NOCS  to estimate object pose and size.
The representation was extended to yield X-NOCS \cite{sridhar2019multiview}, a system capable of reconstructing the visible and occluded surfaces of an object from one or more RGB images. While it can predict camera pose in the canonical NOCS space, it requires depth to estimate metric pose. 
Recently, Chen et al. \cite{chen2020learning} presented the Canonical Shape Space (CASS) that enables category-level object shape, pose and size estimation from an RGB-D image by predicting a pose-independent shape and using it to estimate pose and size.
Both CASS and NOCS are unified object description spaces for shapes in a category. NOCS is explicitly defined by aligning the instances within a category in a normalized 3D space and estimates pose via geometric reasoning. CASS, on the other hand, is implicitly learned via a generative model and enables pose estimation by the network.

Our design is partially inspired by CodeSLAM and its follow-ups \cite{bloesch2018codeslam, zhi2019scenecode, bloesch2019learning}, which use a variational auto-encoder (VAE) to encode the inter-categorical and intra-categorical shape differences in the network. A key difference is that they treat the scene as generic, while we treat it as a collection of specific shapes learned from CAD models (in the form of point clouds).
%which is a common source of supervised training nowadays

% However, during testing, the only information available is an RGB image containing the shape. \PMcomments{SPLIT RGB from RGB-D. These shouldn't be here. CodeSLAM does not estimate the pose of objects since it does not consider objects.}
% RGB-D: avetisyan2019scan2cad,avetisyan2019end,dahnert2019joint, danielczuk2019segmenting
% RGB, no pose since 3d bbox is given: runz2020frodo
% DROPPED: avetisyan2019scan2cad,avetisyan2019end retrieval 
% DROPPED dahnert2019joint no pose, just retrieval. koch2019abc surface normals
% birdal2017 instance reconstruction from multiple point clouds...
% danielczuk2019segmenting segmenting unknwon objects in RGB-D
% kong2017using linear combinations of CAD models

%\PMcomments{Several methods estimate pose implicitly and get low Chamfer distances, but don't evaluate pose explicitly. They are listed in various places.}

We now turn our attention to category-level separate shape and pose estimation from RGB inputs.
The predecessors to our work include approaches that detect 3D objects in RGB images \cite{chen2016monocular,mousavian20173d,xiang2015data} given CAD models from the same class for training. These methods, however, estimate approximate bounding boxes and are limited to objects on the ground plane, with three degrees of freedom.
The IM2CAD system \cite{ izadinia2017im2cad} creates a CAD model of a scene by aligning database CAD models with corresponding objects in the input image.
Georgakis et al. \cite{georgakis2019learning} estimate 6D pose by learning RGB-to-CAD correspondences without requiring photo-realistic texture for the CAD models or explicit 3D pose annotations in the images. 
This is accomplished by learning how to select keypoints and enforcing viewpoint and modality invariance across RGB images and CAD model renderings. % via a quadruplet convolutional neural network.
Several authors \cite{chen2020category,park2020latentfusion} have trained networks on multiple images of known objects  to regress the pose of unseen ones. A similar approach requiring only cluttered images for training  has also been presented by Park et al. \cite{park2020neural}.

% TODO: PM separate RGB from RGB-D inputs 

Assuming the object pose is known or to recover view-dependent shape, probabilistic and generative techniques have been used to impose constraints on 3D shape space. 
A variety of methods are capable of reconstructing 3D shape from a single image given the object pose, or at least an approximation in the form of a bounding box, by implicitly or explicitly learning class-specific generative models. 
Categorized by their shape representations, some use 3D point clouds \cite{fan2017pointsetgeneration,kar2015}, like we do, while others use volumetric representations \cite{choy2016_3d,gwak2017weakly,pinheiro2019domain,tulsiani2017,yan2016perspective,xie2020pix2voxplusplus}, implicit functions \cite{jiang2020sdfdiff,xu2020ladybird}, meshes \cite{Gkioxari_2019_ICCV,henderson2020,kanazawa2018learning,pan2019deep,runz2020frodo,wang2018pixel2mesh,yao2020front2back}, or collections of parametric elements \cite{groueix2018atlasnet}. Other reconstruction approaches from RGB images use the 2.5-D sketch \cite{marrnet}, silhouettes \cite{gwak2017weakly,navaneet2020image}, or a combination of depth maps, silhouettes, 3D voxels, and spherical maps as intermediate representation \cite{genre}. 
AtlasNet \cite{groueix2018atlasnet} can reconstruct a 3D shape, represented as a collection of parametric surface elements, from a single image. High-resolution reconstruction is possible since processing occurs in batches.
Mesh R-CNN \cite{Gkioxari_2019_ICCV}, which we compare with, begins by recognizing objects in a 2D image and infers a coarse 3D occupancy grid, which is converted to a finer-resolution mesh representing the 3D object. Mesh R-CNN, however, approximates the pose from the predicted 3D bounding boxes and does not output precise pose estimates that can be evaluated against ours.
Ye et al. \cite{Ye_2021_CVPR_shelf} present an approach along the same lines, where supervision is provided in the form of foreground masks returned by off-the-shelf object detectors. Shape reconstruction is carried out in two phases: category-level volumetric prediction and instance-level mesh specialization. Pose supervision is not required, and the accuracy of pose estimation is not assessed.

Our work is also relevant to object pose estimation.
We refer readers to \cite{sahin2020review} for a survey and to \cite{Hodan_2018_ECCV} for a benchmark and a comprehensive evaluation of several methods.
Several authors have developed methods relying on RGB-D inputs \cite{avetisyan2019scan2cad,chen2020g2l,choi20123d,avetisyan2019end,hodavn2015detection,kehl2016deep,tejani2018latent,wang2020six,wang2019densefusion} to overcome clutter and occlusion.
Clearly, pose estimation is harder based solely on RGB information, even when exact CAD models are used for training. Methods in this category can be further distinguished into those that learn to estimate 2D-3D correspondences and feed them to a perspective-n-point (PnP) solver \cite{brachmann2016uncertainty,grabner2019gp2c,li2019cdpn,park2019pix2pose,peng2019pvnet}, regress pose from image features \cite{chen2020g2l,xiang2018posecnn}, perform retrieval and/or refinement \cite{song2020hybridpose,sundermeyer2020multi}, or estimate the projections of the corners of the bounding box in the image \cite{kehl2017ssd,tekin2018real}. 
Li et al. \cite{li2018deepim} estimate pose at the category level by iterative rendering and refinement.

\section{Method: Glissando-Net} \label{sec:method}
%We introduce the Glissando-Net, for single view category level pose estimation and 3D shape (in the form of a point cloud) reconstruction. 
Figure \ref{fig:workflow} shows an overview of the structure of our proposed Glissando-Net. 

\noindent\textbf{Training Stage} (delineated by the solid lines in Fig. \ref{fig:workflow}~(a)). The model learns to predict the pose and canonical shape in a supervised manner. The inputs to the network are (1) the object's RGB image, (2) 3D shape, %represented as a 
in the form of a
3D point cloud, and (3) its ground truth pose with respect to the RGB view. The \emph{canonical shape} (e.g. the bottom-left point cloud of a camera in Fig. \ref{fig:workflow} (a)) is defined as in \cite{wang2019normalized}: by aligning all objects in the same category to the same pose according to some convention (e.g. axis-aligned and frontal) and normalized to the same scale. % in a unit space. 
For 2D appearance learning, we use a U-Net-based encoder-decoder network \cite{RFB15a} (top branch in Fig. \ref{fig:workflow} (a)). 
%to learn the RGB image feature. 
For 3D shape learning, we use a Variational Auto Encoder (VAE) network (bottom branch in Fig. \ref{fig:workflow} (a)) to encode the shape information from the canonical point cloud. We define the latent code as $\mathbf{z}$. The ground truth pose is used to regress the predicted pose, as well as to project the 3D points to the 2D image space, such that the geometrically correct features learned from the 2D appearance branch can be passed to the 3D shape inference branch.

\noindent\textbf{Inference Stage} (delineated by the dashed lines in Fig. \ref{fig:workflow}~(a)). We predict an object's canonical 3D point cloud and its pose, given \emph{only} one RGB image as input. Since no 3D shape or pose is given, we simply use all-zero-code for the $\mathbf{z}$ vector in the VAE, combined with the RGB information passed to the shape decoder part of the VAE to predict the final shape and pose. Such an ill-posed task is made possible because shape reconstruction is primarily guided by the RGB appearance passed from the top U-Net, while the bottom point cloud VAE ``refines" the shape based on the learned shape-appearance priors in the training stage.

\subsection{Image Feature Extraction} \label{subsec:method_image}
The RGB sub-network of Glissando-Net (top in Fig. \ref{fig:workflow}~(a)) is a U-Net \cite{RFB15a} comprising an encoder and a decoder. The encoder follows a typical convolutional architecture with a stack of five groups of layers. Each group performs two 3$\times $3 convolutions, each followed by a group normalization layer \cite{Wu_2018_ECCV}  and a ReLU. The first 3$\times $3 convolution is used for downsampling with a stride of 2 and doubling the number of feature channels. We use group normalization instead of batch normalization, because the computation of group normalization is independent of batch sizes, and its accuracy is stable in a wide range of batch sizes. We also stack five groups of layers in the decoder. Each group consists of a 2$\times $2 deconvolution and two 3$\times $3 convolutions, each followed by a group normalization layer and a ReLU. 2$\times $2 deconvolution is used for upsampling, while the number of channels is halved. Layers in the decoder are skip-connected to the corresponding layers in the encoder. 

%Each deconvolution is followed by a feature concatenation with the feature map from the corresponding layer in the encoder. 

%The skip connections between feature maps from encoder and decoder help our network utilize the RGB image features.

\subsection{Geometric Feature Extraction} \label{subsec:method_shape}
The geometric sub-network of Glissando-Net (bottom network in Fig. \ref{fig:workflow} (a)) is a Variational Auto Encoder (VAE), which takes as input a point cloud sampled from a CAD model. All objects are aligned to the same canonical pose.
%PM: R=I, t=0 doesn't really mean anything.
%with identity rotation and zero translation. 
An example is shown in the bottom-left of Fig. \ref{fig:workflow} (a): the frontal point cloud of a camera. %point-and-shoot camera.

The backbone of the encoder is a PointNet++ \cite{qi2017pointnet++} with five set abstraction layers. PointNet++ is a hierarchical neural network that applies PointNet recursively on a nested partitioning of the input 3D point set. It partitions the set of points into overlapping local regions according to L1 distance in the underlying space and it learns local features in increasing contextual scales. 
% This PointNet++ based encoder extract
% geometric feature from the canonical point cloud at different scales. To encode the geometry information of the point cloud, we apply the VAE with a variational reparameterization process in the bottleneck. It 
This PointNet++ based encoder transforms the geometric features from the canonical point cloud into a latent vector $\mathbf{z}$ following a normal distribution. %Using variational  reparameterization -- PM: term does not appear in Kingma paper
After reparameterization, we have two fully connected layers to estimate its mean $\mu$ and variance $\sigma^2$. The latent code $\mathbf{z}$ is then sampled from a Gaussian distribution $N(\mu,\sigma^2)$. KL-divergence is used to constrain the code distribution close to a centered isotropic Gaussian distribution. The decoder part consists of five fully-connected layers (with 256, 512, 1024, 2048, and 4096 nodes respectively), which generate an abstract representation for the point cloud in different scales. Group normalization and ReLU are applied between  layers. 

\subsection{Feature Transform Module} \label{subsec:method_concate}
The top network (image encoder-decoder) extracts RGB features from the 2D image and the bottom network (point cloud VAE) extracts geometric features from the 3D point cloud. In order to fully utilize the information from both, we transform and concatenate the image decoder features with the point cloud encoder and the point cloud decoder.

As shown in Fig.~\ref{fig:workflow} (b), to concatenate the image decoder features with those of the point cloud encoder, we project the 3D points to the 2D image %using intrinsic matrix and the 3D pose between the camera view and the canonical 3D shape. 
according to the relative pose between the camera and the canonical 3D shape. 
We denote the sampled 3D points in the $i^{th}$ scale (abstraction layer) in the point cloud encoder by $M_{i}$. Both our point cloud encoder and the image decoder have five scales 
%and the image decoder has the same number of scales. So we project the 3D points in different scale 
allowing us to project $M_{i}$ 
to the corresponding feature map of the image decoder. $F_{i}$ denotes the feature map in the $i^{th}$ scale of the image decoder.  We use the ROI-align 
layer from mask-RCNN \cite{he2017mask} to obtain the local features around the projected points. Then, we concatenate the features from the ROI-align layer with the corresponding 3D point features from the point cloud encoder $F_P(M_{i})$ :

\begin{small}
\begin{equation}
     % Feature3D_{i} --- PM: it's not used later anyway and it looks bad
     %F3D_i = Concatenate(RoI_{align}(F_{5-i}(K \cdot (R \cdot M_{i} + T))), FP(M_{i}))
     F_{3D_i} = Conc(RoI_{align}(F_{5-i}(K \cdot (R \cdot M_{i} + T))), F_P(M_{i}))
\end{equation}
\end{small}

\noindent where the ground truth rotation and translation are denoted by $R$ and $T$ and the intrinsic matrix by $K$. 
This technique creates strong connections between each 3D point and the corresponding 2D pixel. % from image. 

The point cloud encoder is only used during training, since there is no 3D shape input during inference. To concatenate image decoder features with those from the point cloud decoder, %we cannot project the 3D points to 2D image. Instead 
we transform the features from the entire image map at scale $i$ ($F_{i}$) into a feature vector using a convolutional layer and a fully-connected layer. We denote the feature vector from the $i^{th}$ scale of the point cloud decoder by $F_{D_{i}}$. Then, we concatenate $F_{i}$ with the feature vector of the corresponding layer from the point cloud decoder ($F_{D_{5-i}}$). 

% Based on these two kinds of feature transformation, the point cloud VAE is able to encode the information from both 2D image and 3D point cloud. Then our network can regress to 3D shape and pose regression from both 2D and 3D feature space.

With these two feature transformations, we facilitate interaction between 2D and 3D features. In training, the prediction from the point cloud decoder is partially conditioned on the 2D features, so that we can estimate pose and reconstruct 3D shape in testing without point cloud inputs.

%------------------------------------------------------------------------
\begin{figure*}[thb]
    \begin{center}
        \includegraphics[width=1.0\textwidth]{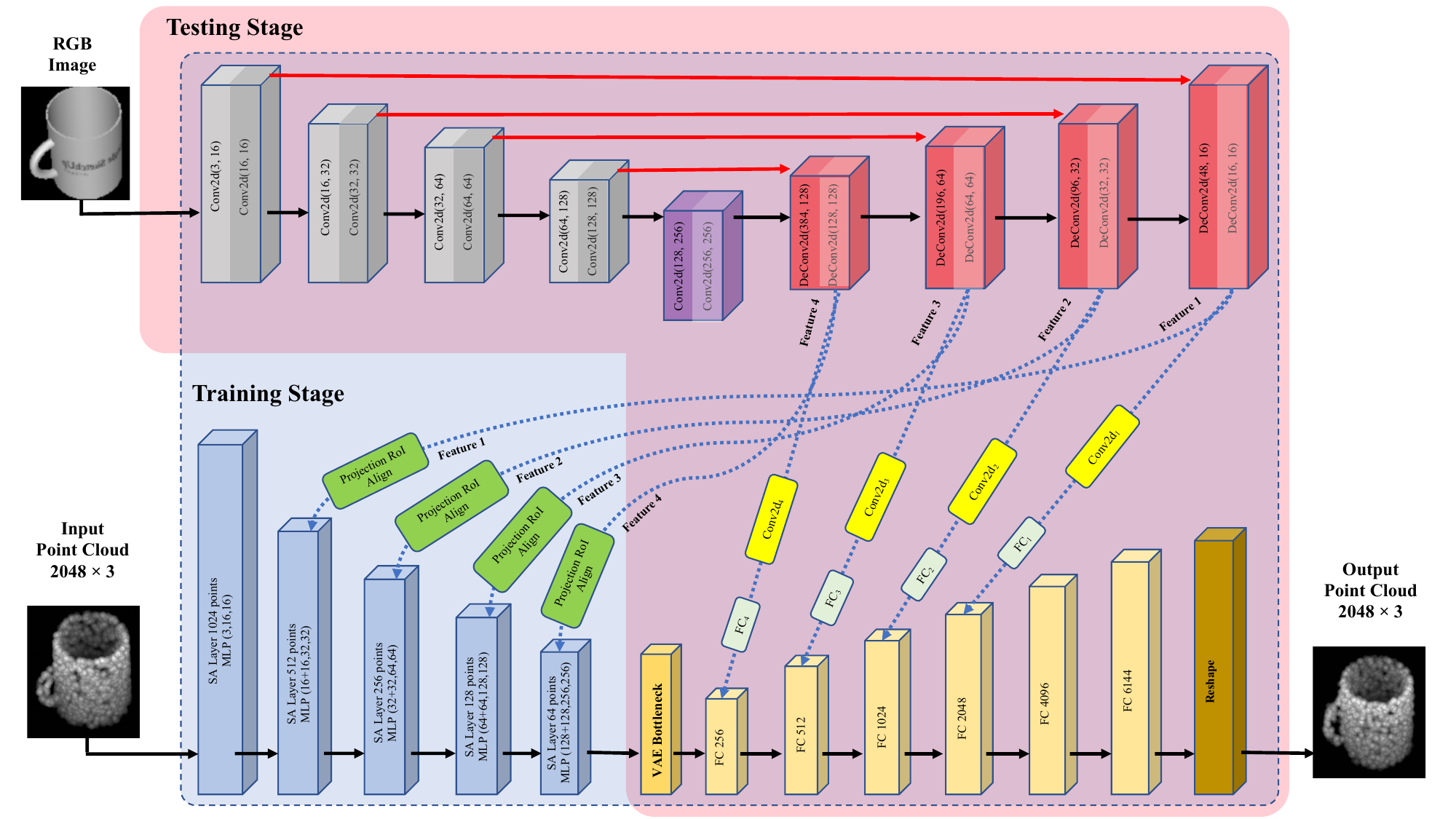}
    %\vspace{-2pt}
    \caption{Details of RGB and point cloud encoder-decoder from Glissando-Net. Red arrows indicate skip connections between RGB encoder and decoder layers. Features 1–4 from the RGB decoder are fed into the point cloud encoder-decoder. The point cloud encoder uses set abstraction (SA) layers with sampled points from PointNet++, which are projected back to the RGB feature map using RoI align. Features 1–4 are concatenated with the point cloud decoder outputs via 2D convolution and fully connected layers.}\label{fig:network_details}
    \end{center}
\end{figure*}

\subsection{Details of Network Structure}
%\subsection{RGB and Point Cloud Encoder-Decoders}

Figure \ref{fig:network_details} shows the details of the RGB and point cloud encoder-decoders, zooming in on parts of Fig. \ref{fig:workflow}. $Conv2d(n_{in},n_{out})$ denotes a $3\times3$ 2D convolution layer with $n_{in}$ input channels and $n_{out}$ output channels. $DeConv2d(n_{in},n_{out})$ denotes a $3\times3$ 2D deconvolution layer with $n_{in}$ input channels and $n_{out}$ output channels. The red arrows denote the skip connections between the encoder layer and the decoder layer. Feature 1 - Feature 4 are the outputs from the RGB decoder layers. We feed these features to the corresponding layers of point cloud encoder-decoder. In the point cloud encoder, $SA\;Layer\;n\;points$ denotes a set abstraction layer with $n$ sampled points from PointNet++ \cite{qi2017pointnet++}. We project the sampled points in the SA layers back to the corresponding feature map of the  RGB encoder. Then, we use RoI align to take the output features from the RGB encoder-decoder Feature 1 - Feature 4 and concatenate them with the output from the corresponding SA layers. To concatenate Feature 1 - Feature 4 with the output of the point cloud decoder layers, we use one 2D convolution layer and one fully connected layer to convert each of Feature 1 - Feature 4 to a one-dimensional feature.

%------------------------------------------------------------------------

\subsection{3D Shape and Pose Regression} \label{subsec:method_pose}
We use one fully-connected layer after the point cloud decoder to regress the 3D shape. We use the Earth Mover's Distance (EMD) or the Chamfer Distance as the shape reconstruction loss \cite{fan2017pointsetgeneration}. 
EMD is computationally more expensive since it solves an assignment problem, while Chamfer distance is faster to compute but leads to a slight loss of accuracy as shown in Section~\ref{sec:results}.

We design two small networks, each with five fully connected layers for pose estimation. These two networks are of the same structure up to the last layer: one outputs a 4D quaternion for the rotation, and the other outputs a 3D translation. We design a loss function named Per-Point L2 loss for pose regression. We, first, apply the estimated rotation $\hat{R}$ and translation $\hat{T}$ to input point cloud $pc_{gt}$ to obtain $pc_{1}$, and the ground truth rotation $R$ and translation $T$ to input point cloud $pc_{gt}$ to obtain $pc_{2}$. 
Then, we define the Per-Point L2 loss between these point clouds, $pc_{1}$ and $pc_{2}$, as: %Per-Point L2 loss $Loss_{per-point}$:
\begin{equation}
    Loss_{per\text{-}point} = \frac{1}{N} \sum_{x \in pc_{gt}} \left\|(\hat{R}x + \hat{T}) - (Rx + T) \right\|_2
\end{equation}
\noindent where $N$ is the number of points in $pc_{1}$ and $pc_{2}$.

Our total loss is the sum of three terms: shape reconstruction loss $Loss_{shape}$, KL Divergence loss $Loss_{kld}$ and pose loss $Loss_{per\text{-}point}$. We use weights $w_{shape}$, $w_{kld}$ and $w_{pose}$ to balance the terms.
\begin{small}
\begin{equation}
    Loss_{all} = w_{shape} Loss_{shape} + w_{kld} Loss_{kld} + w_{pose} Loss_{per\text{-}point} \label{eq:loss}
\end{equation}
\end{small}

%------------------------------------------------------------------------
\section{Experimental Results} \label{sec:results}
\newcommand\widf{.40\columnwidth}

In this section, we present the datasets and metrics used in our experiments, experimental results, as well as several ablation studies.

% \PMcomments{From rebuttal: ``Table 4 should report mean Chamfer distance
% (CD) over objects within a category, and then mean CD
% over categories. As R3 points out, we inadvertently reported
% the wrong means for our method. The correct means are: Glissando-Net(S1) 5.06, Mesh RCNN (S2) 17.59 (over
% 7 categories), Glissando-Net(S2) 14.39 (over 9 categories) and 11.83 (over 7 categories). In Table 1, the correct means are: Glissando-Net(Chamfer) 0.91, Glissando-Net(EMD)
% 2.62, Glissando-Net(EMD) 0.85. In both cases, the correct mean errors of our method are still much lower than those of other methods. This correction will be applied immediately." }

\subsection{Datasets, Experimental Setup and Evaluation Metrics.} \label{subsec:experi_data}
We evaluate Glissando-Net on two recent datasets for category-level pose estimation and shape reconstruction. (1) The \textbf{NOCS dataset} \cite{wang2019normalized} consists of six object categories: bottle, bowl, camera, can, laptop and mug, picked from ShapeNetCore \cite{chang2015shapenet}. NOCS includes both synthetic and real data with camera intrinsics and ground truth pose. The synthetic dataset consists of 275K training and 25K validation images rendered from 1085 individual object instances in ShapeNetCore. The real scene data consists of 5250 training images and 2750 testing images from 18 different real scenes. We crop the objects in each image using the ground truth mask and resize to 128 $\times$ 128. For synthetic scenes, we generate about 625K training crops and about 46K validation crops. For real scenes, we have about 26K training and 9K testing images. We sample point clouds with 2048 points from the CAD models for both data types. %NOCS dataset provides intrinsics and ground truth pose. 
(2) The \textbf{Pix3D dataset} \cite{sun2018pix3d} has 395 3D models of furniture from 9 categories, with 10,069 precisely aligned 
RGB images, %It has RGB images and ground truth shapes with precise 2D-3D alignment. This dataset has 
camera intrinsics and ground truth pose. We crop the object in each image using the ground truth mask and resize it to 256 $\times$ 192. %For a fair comparison, 
To be consistent with  \cite{nie2020total3dunderstanding, Gkioxari_2019_ICCV} which we compare with in the experiments, we sample point clouds with 10K points from the CAD models. We evaluate the 3D shape reconstruction on the Pix3D dataset following the train/test splits from Gkioxari et al.~\cite{Gkioxari_2019_ICCV}. 
(3) The \textbf{Objectron dataset} \cite{objectron2021} is a collection of short, object-centric video clips, which are accompanied by AR session metadata that includes camera poses, sparse point-clouds and characterization of the planar surfaces in the surrounding environment. In each video, the camera moves around the object, capturing it from different angles. The dataset consists of 15K annotated video clips supplemented with over 4M annotated images in the following categories: bikes, books, bottles, cameras, cereal boxes, chairs, cups, laptops, and shoes.

\begin{table*}[thb]
    \caption{Shape Reconstruction Result on Synthetic and Real Data from NOCS. CASS requires RGB-D inputs. Glissando-Net (EMD)$^*$ represents Glissando-Net (EMD) without the 3D encoder.} \label{table_nocs_shape}
	\begin{center}
	    %\scalebox{0.8}{
    	\begin{tabular}{|c|c|c|c|c|c|c|c||c|c|c|}
			\hline
			\multirow{3} * {Methods} &
			\multicolumn{10}{|c|}{Mean Chamfer Distance in mm $\downarrow$} \\
			\cline{2-11}
			& \multicolumn{7}{|c||}{Real Data} & \multicolumn{3}{|c|}{Synthetic Data} \\
			\cline{2-11}
			& bottle & bowl & camera & can & laptop & cup & overall & cup & bottle & camera \\ 
			\hline
			CPS \cite{manhardt2020cps} & - & - & - & - & - & - & - & 3.4 & 3.7 & 11.2  \\
			\hline
			CASS \cite{chen2020learning} & 0.75 & 0.38 & \textbf{0.77} & 0.42 & 3.73 & 0.32 & 1.06 & - & - & -  \\
            \hline
            OLD-Net \cite{fan2022object} & 2.47 & 1.74 & 2.15 & 2.05 & 2.03 & 2.03 & - & 1.78 & 2.25 & 1.92  \\
            \hline
            NOCE \cite{lee2021category} & - & - & - & - & - & - & 31.44 & - & - & - \\
			\hline
			%Glissando-Net (Chamfer) & 0.49 & 0.75 & 1.81 & 0.25 & 1.87 & 0.33 & 0.95 & 0.54 & 0.89 & 1.31 \\
			%\hline
			%Glissando-Net (EMD) & \textbf{0.36} & 0.64 & 1.75 & \textbf{0.24} & \textbf{1.28} & 0.35 & 0.74 & \textbf{0.39} & \textbf{0.63} & 1.16 \\
			%\hline
			%\multicolumn{11}{|c|}{Train on Synthetic + Real} \\
			\hline
			Glissando-Net (Chamfer) & \textbf{0.42} & 0.47 & 1.21 & 0.35 & 2.62 & 0.36 & 0.91 & 0.41 & 1.22 & 0.75 \\
			\hline
			Glissando-Net (EMD)$^*$ & 2.72 & 1.88 & 2.69 & 2.91 & 4.07 & 1.43 & 2.62 & 2.15 & 2.78 & 5.33 \\
			\hline
			Glissando-Net (EMD) & 0.49 & \textbf{0.34} & 1.09 & \textbf{0.34} & \textbf{2.54} & \textbf{0.31} & \textbf{0.85} & \textbf{0.39} & \textbf{0.81} & \textbf{0.69} \\
			\hline
			
		\end{tabular}
		%}
	\end{center}
	%\vspace{-8pt}
\end{table*}

\begin{table*}[thb]
    \caption{Pose Estimation Result on Synthetic and Real Data from NOCS. Glissando-Net (EMD)$^*$ represents Glissando-Net (EMD) without the 3D encoder.}	\label{table_nocs_pose}
	\begin{center}
    	%\scalebox{0.8}{
    	\begin{tabular}{|c|c|c|c|c|c|c|}
    		\hline
    		\multirow{2} * {Methods} &
    		\multicolumn{3}{|c|}{APP ($\alpha$ = 0.2/0.5) $\uparrow$} &
    		\multicolumn{3}{|c|}{10\degree \& 10 cm $\uparrow$} \\
    		\cline{2-7}
    		&cup & bottle & camera & cup & bottle & camera \\ 
    		\hline
    		\multicolumn{7}{|c|}{Synthetic Data} \\
    		%\hline
    		%No Weighting & 17.2\%/49.2\% & 16.1\%/48.7\% & 10.5\%/35.4\% & 11.1\% & 19.1\% & 8.5\% \\
    		%\hline
    		%Multi-Task Weighting \cite{kendall2018multi} & 19.6\%/50.1\% & 14.2\%/47.8\% & 13.4\%/35.2\% & 27.8\% & 20.8\% & 9.6\% \\
    		\hline
    		CPS \cite{manhardt2020cps} & 21.6\%/52.9\% & 18.8\%/56.5\% & 14.8\%/\textbf{43.7\%} & 28.8\% & 25.6\% & 12.5\% \\
    		%\hline
    		%Glissando-Net (Chamfer) & 45.51\%/64.23\% & 38.64\%/59.77\% & 13.08\%/33.07\% & 34.51\% & 22.77\% & 7.98\% \\
    		%\hline
    		%Glissando-Net (EMD) & 52.85\%/70.13\% & 43.50\%/65.41\% & \textbf{17.34\%}/34.89\% & 41.54\% & \textbf{25.97\%} & 11.15\% \\
    		%\hline
    		%\multicolumn{7}{|c|}{Train on (Synthetic + Real) \& Test on Synthetic} \\
    		\hline
    		Glissando-Net (Chamfer) & 46.12\%/63.86\% & 34.02\%/59.12\% & 15.95\%/34.23\% & 34.55\% & 23.09\% & 8.85\% \\
    		\hline
    		Glissando-Net (EMD)$^*$ & 34.17\%/53.23\% & 30.84\%/51.15\% & 13.41\%/28.47\% & 21.12\% & 19.25\% & 5.57\% \\
    		\hline
    		Glissando-Net (EMD) & \textbf{56.73\%/73.57\%} & \textbf{50.70\%/71.08\%} & \textbf{16.05}\%/32.94\% & \textbf{44.65\%} & \textbf{25.99}\% & \textbf{12.62\%} \\
    		\hline
    		\multicolumn{7}{|c|}{Real Data} \\
    		\hline
    		CPS \cite{manhardt2020cps} & \textbf{36.8}\%/64.6\% & 1.6\%/14.1\% & 19.6\%/\textbf{56.7}\% & 1.3\% & 9.5\% & 1.9\% \\
    		%\hline
    		%Glissando-Net (Chamfer) & 38.84\%/68.03\% & 3.44\%/15.25\% & 19.82\%/45.74\% & 2.64\% & 7.04\% & 1.49\% \\
    		%\hline
    		%Glissando-Net (EMD) & \textbf{39.44\%/81.07\%} & \textbf{9.95\%/27.59\%} & 22.96\%/52.34\% & 2.71\% & \textbf{11.29\%} & 2.23\% \\
    		%\hline
    		%\multicolumn{7}{|c|}{Train on (Synthetic + Real) \& Test on Real Data} \\
    		\hline
    		Glissando-Net (Chamfer) & 31.09\%/74.39\% & 5.73\%/19.19\% & 20.45\%/44.79\% & 3.87\% & 8.62\% & 5.78\% \\
    		\hline
    		Glissando-Net (EMD)$^*$ & 18.78\%/58.43\% & 3.67\%/18.32\% & 6.67\%/18.32\% & 1.08\% & 2.33\% & 1.33\% \\
    		\hline
    		Glissando-Net (EMD) & 34.48\%/\textbf{78.99}\% & \textbf{9.28}\%/\textbf{22.80}\% & \textbf{24.91\%}/42.80\% & \textbf{4.44\%} & \textbf{10.16}\% & \textbf{11.36\%} \\
    		\hline
    	\end{tabular}
    \end{center}
    %\vspace{-8pt}
	%}
\end{table*}

\begin{table*}[thb]
    \caption{Overall Pose Estimation Result on Synthetic and Real Data from NOCS. Glissando-Net (EMD)$^*$ represents Glissando-Net (EMD) without the 3D encoder.}\label{table_nocs_pose_all}
	\begin{center}
	    %\scalebox{0.9}{
    	\begin{tabular}{|c|c|c|c|c|}
			\hline
			\multirow{2} * {Methods} &
			\multicolumn{2}{|c|}{Synthetic Data} & \multicolumn{2}{|c|}{Real Data} \\
			\cline{2-5}
			& 10\degree \& 10 cm $\uparrow$ & APP ($\alpha$ = 0.2/0.5) $\uparrow$ & 10\degree \& 10 cm $\uparrow$ & APP ($\alpha$ = 0.2/0.5) $\uparrow$ \\ 
			\hline
			CPS \cite{manhardt2020cps} & 31.7\% & 19.1\%/49.6\% & 14.7\% & 30.8\%/64.0\% \\
			\hline
			CPS ++ \cite{manhardt2020cpsv3} & 27.4\% & 17.8\%/45.2\% & \textbf{22.3\%} & \textbf{41.0\%/73.6\%} \\
			\hline
            OLD-Net \cite{fan2022object} & 23.4\% & - & 9.8\% & - \\
            \hline
            NOCE \cite{lee2021category} & 19.2\% & - & 9.6\% & - \\
            \hline
			Glissando-Net (Chamfer) & 31.22\% & 34.25\%/54.02\% & 16.27\% & 30.99\%/61.98\% \\
			\hline
			Glissando-Net (EMD)$^*$ & 21.21\% & 30.55\%/49.47\% & 2.97\% & 23.78\%/37.82\% \\
			\hline
			Glissando-Net (EMD) & \textbf{31.86\%} & \textbf{47.11\%/66.65\%} & 19.08\% & 34.06\%/63.90\% \\
			\hline
		\end{tabular}
		%}
	\end{center}
	%\vspace{-8pt}
\end{table*}

%\subsection{Experimental Setup} \label{subsec:experi_setup}
We implemented our method in PyTorch. We train our network, which has 67 million parameters, on all images and point clouds generated both from the synthetic and real training data of NOCS with a batch size of 128 for 180 epochs using the ADAM optimizer. We use an initial learning rate of 0.0001 and decay it by 0.1 after 100, 130, and 160 epochs. Training on 4 NVIDIA GTX 2080Ti GPU takes approximately 10 days. We test our model on the synthetic and real data from NOCS separately with inference taking 0.314 seconds per frame. 
%To evaluate our model on the real data of NOCS, we finetune the last two layers of the network on the real training data for another 300 epochs.To make our model more general, we also train our network on both synthetic and real data, then test on synthetic and real data separately.
%
For the Pix3D dataset, we train our network with a batch size of 32 for 160 epochs using the ADAM optimizer. We also use an initial learning rate of 0.0001, decayed after 110 and 150 epochs by 0.5 and 0.2. We set the weights $w_{shape}$, $w_{kld}$ and $w_{pose}$ in Eq. (\ref{eq:loss}) to 1, 100 and 100, respectively.
%\subsection{Evaluation Metrics} %\label{subsec:experi_metric}

\noindent\textbf{3D Shape Evaluation} We use Chamfer distance %defined by Equation~ \ref{chamfer_formula} 
to evaluate 3D shape reconstruction. 

\noindent\textbf{Pose Estimation} We use two metrics to evaluate pose estimation: the \emph{Average Distance of Predicted Point Sets (APP)}~\cite{manhardt2020cps}  and the \emph{10\degree \& 10 cm metric} \cite{wang2019normalized}. APP is extended from two %of the most common 
metrics for 6D pose comparison known as the Average Distance of Distinguishable Model Points (ADD) and Average Distance of Indistinguishable Model Points (ADI)~\cite{hodavn2016evaluation,hinterstoisser2011multimodal}. ADD measures whether the average point distance between the point cloud in the estimated 
%pose 
and 
%the point cloud in 
the ground truth pose is less than 10\% of the object’s diameter. ADI extends ADD for symmetries, measuring error as the mean distance to the closest model point. To circumvent the need for direct correspondences, 
%and be agnostic to scale discrepancies, 
Manhardt et al.~\cite{manhardt2020cps} introduce APP which simply extends ADI to be computed bidirectionally: 
\begin{equation}
    APP = \left\{ 
    \begin{array}{cc}
         1, & \mbox{if } m_{1} \leq \alpha \cdot d(pc) \land m_{2} \leq \alpha \cdot d(\hat{pc}) \\
         0, & otherwise
    \end{array}
    \right.
\end{equation}
$pc$ is the point cloud with ground truth pose, $\hat{pc}$ is the point cloud with estimated pose. $d(\cdot)$ measures the diameter of $pc$. $m_{1}$ and $m_{2}$ are defined as:
\begin{equation}
\begin{split}
    m_{1} = \frac{1}{N} \sum_{x_{1} \in pc} \min_{x_{2} \in \hat{pc}} \left\|(Rx_{1} + T) - (\hat{R}x_{2} + \hat{T})\right\|_2,\\
    m_{2} = \frac{1}{N} \sum_{x_{2} \in \hat{pc}} \min_{x_{1} \in pc} \left\|(Rx_{1} + T) - (\hat{R}x_{2} + \hat{T})\right\|_2
\end{split}
\end{equation}
$N$ is the number of points in $pc$. $R$ and $T$ are the ground truth rotation and translation, and $\hat{R}$ and $\hat{T}$ are the estimated rotation and translation. We use 20\% and 50\% as threshold values for $\alpha$. 
We also use the 10\degree \& 10 cm metric, which counts the number of samples for which rotation error is less than 10\degree and translation error is less than 10 cm.

\subsection{NOCS Dataset Results} \label{subsec:experi_nocs}
%We evaluate Glissando-Net on synthetic and real data from NOCS dataset for 3D shape reconstruction and pose estimation.
We compare the performance of Glissando-Net with CPS \cite{manhardt2020cps}, CASS \cite{chen2020learning}, OLD-Net \cite{fan2022object} and NOCE \cite{lee2021category}. To the best of our knowledge, CPS is the only method for 3D shape reconstruction and category level pose estimation on real data using an RGB image as input. We evaluate our method on the three categories which CPS reports, even though our model was trained on all categories, using both real and synthetic data. CASS requires RGB-D inputs and NOCE requires some depth information as input.

%We show results from Glissando-Net trained on both real and synthetic data, since it performs better than training on only the latter due to the increased diversity of the training set.

\textbf{Glissando-Net without Point Cloud Encoder (Glissando-Net$^*$):} We remove the point cloud input and 3D encoder to create an additional baseline. We compare the performance of Glissando-Net with this baseline to evaluate the importance of the point cloud encoder.
% We observe that by having 3D inputs in training the learned model fits training data better and generalize better.
\begin{table*}[thb]
	%\vspace{-14pt}
	\caption{Shape Reconstruction Results on Pix3D. The values are in units of 10$^{-3}$. (AtlasNet and TMN results are repeated from Total3D.)}\label{table_pix3d_instance_shape}
	\begin{center}
	    %\scalebox{0.82}{
		\begin{tabular}{|c|c|c|c|c|c|c|c|c|c|c|}
			\hline
			Methods & bed & bookcase & chair & desk & sofa & table & tool & wardrobe & misc & mean \\ 
			\hline
			\multicolumn{11}{|c|}{Instance-level Shape Reconstruction} \\
			\hline
			AtlasNet \cite{groueix2018atlasnet} ($S_{1}$) & 9.03 & 6.91 & 8.37 & 8.59 & 6.24 & 19.46 & 6.95 & 4.78 & 40.05 & 12.26 \\
			\hline
			TMN \cite{pan2019deep} ($S_{1}$) & 7.78 & 5.93 & 6.86 & 7.08 & 4.25 & 17.42 & 4.13 & 4.09 & 23.68 & 9.03 \\
			\hline
			Total3D \cite{nie2020total3dunderstanding} ($S_{1}$) & 5.99 & 6.56 & 5.32 & 5.93 & 3.36 & 14.19 & \textbf{3.12} & 3.83 & 26.93 & 8.36 \\
			\hline
			Glissando-Net ($S_{1}$) & \textbf{2.61} & \textbf{4.85} & \textbf{2.21} & \textbf{4.00} & \textbf{1.28} & \textbf{3.75} & 13.39 & \textbf{1.82} & \textbf{11.68} & \textbf{5.06} \\
			\hline
			%\hhline{|=|=|=|=|=|=|=|=|=|=|=|}
			\multicolumn{11}{|c|}{Category-level Shape Reconstruction} \\
			\hline
			Mesh R-CNN \cite{Gkioxari_2019_ICCV} ($S_{2}$) & 15.48 & \textbf{14.14} & 9.90 & 20.70 & 5.30 & 26.54 & - & - & 31.08 & 17.59 \\
			\hline
			Glissando-Net ($S_{2}$) & \textbf{5.98} & 14.22 & \textbf{6.65} & \textbf{15.51} & \textbf{1.69} & \textbf{13.99} & 40.96 & 5.78 & \textbf{24.78} & \textbf{11.83} \\
			\hline
		\end{tabular}
		%}
	\end{center}
	%\vspace{-8pt}
\end{table*}

% \begin{table}[thb]
% 	\begin{center}
% 	    \scalebox{0.83}{
% 		\begin{tabular}{|c|c|c|c|c|}
% 			\hline
% 			\tabincell{c}{Code \\ Size} & \tabincell{c}{Shape \\ Chamfer mm $\downarrow$} & \tabincell{c}{APP \\ ($\alpha$ = 0.5) $\uparrow$} & \tabincell{c}{APP \\ ($\alpha$ = 0.2) $\uparrow$} & \tabincell{c}{10\degree \\ 10cm $\uparrow$} \\ 
% 			\hline
% 			32 & 0.98 & 52.50\% & 28.50\% & 6.67\% \\
% 			\hline
% 			64 & 0.90 & 57.67\% & 30.83\% & 14.50\% \\
% 			\hline
% 			128 & 0.87 & 57.33\% & 31.17\% & 16.83\% \\
% 			\hline
% 			256 & 0.81 & 58.50\% & 35.50\% & 18.50\% \\
% 			\hline
% 			512 & 0.81 & 58.83\% & 34.50\% & 17.91\% \\
% 			\hline
% 			1024 & 0.80 & 58.67\% & 35.70\% & 18.84\% \\
% 			\hline
% 		\end{tabular}
% 		}
% 	\end{center}
% 	\vspace{-8pt}
% 	\caption{Ablation study on code size}\label{table_nocs_ab_code}
% \end{table}

%\begin{figure}[thb]
    %\centering
    %\includegraphics[width=0.67\linewidth]{figure/Vis2New.pdf}
    %\vspace{-16pt}
    %\caption{Qualitative results on NOCS dataset: the ground-truth 3D shape with ground-truth rotation and our predicted 3D shape with predicted rotation.}
    %\label{fig:visNOCs}
%\end{figure}

\begin{figure*}[thb]
\begin{center}
\scalebox{0.9}{
\begin{tabular}{ccccc}
\includegraphics[width=\widf]{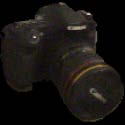} & \includegraphics[width=\widf]{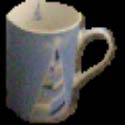} & \includegraphics[width=\widf]{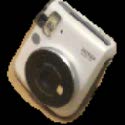} & \includegraphics[width=\widf]{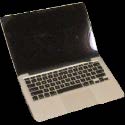} & \includegraphics[width=\widf]{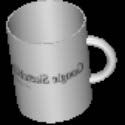} \\
\includegraphics[width=\widf]{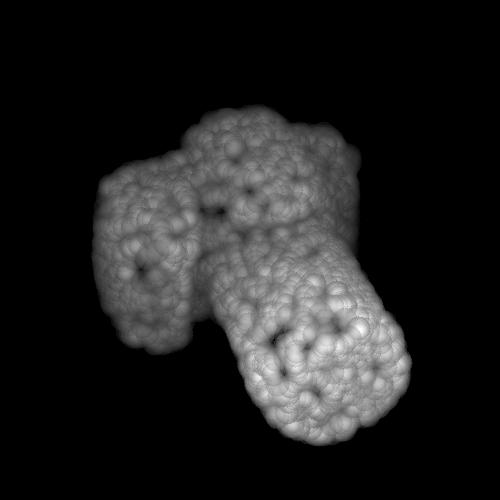} & \includegraphics[width=\widf]{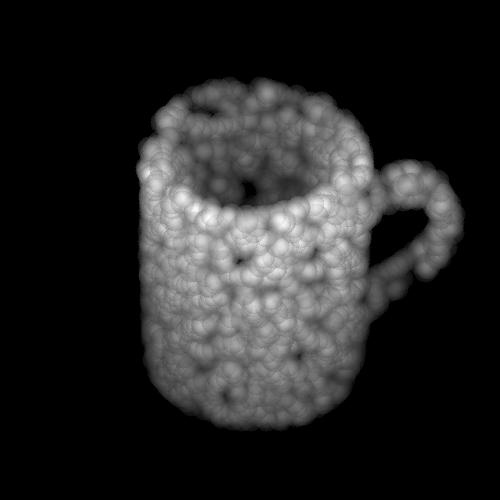} & \includegraphics[width=\widf]{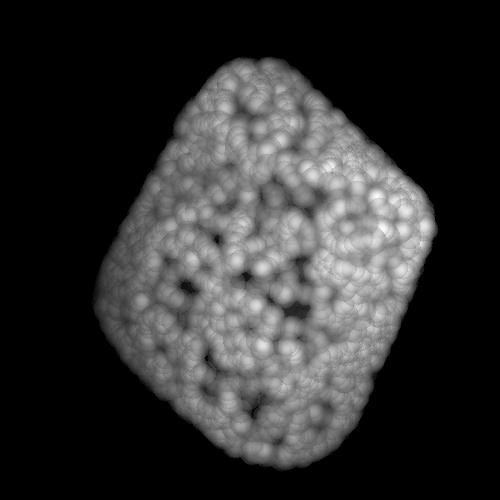} & \includegraphics[width=\widf]{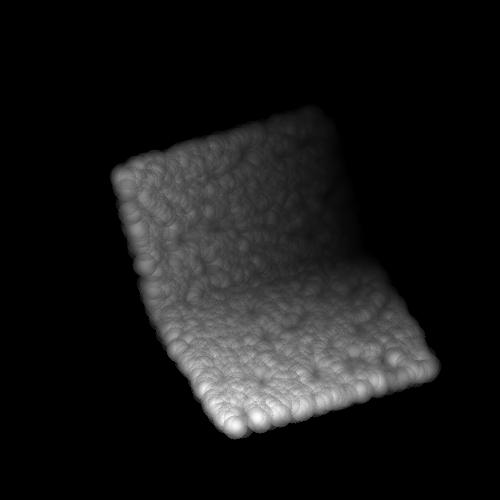} & \includegraphics[width=\widf]{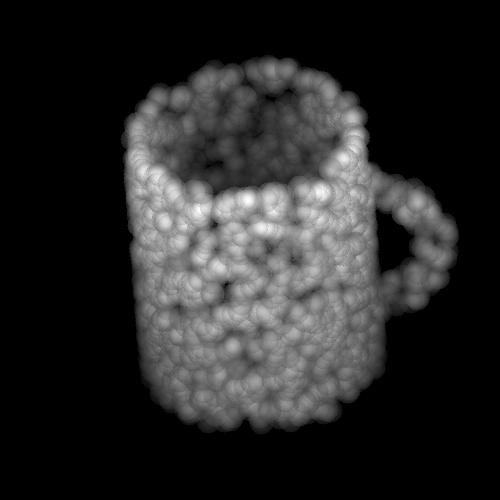} \\
\includegraphics[width=\widf]{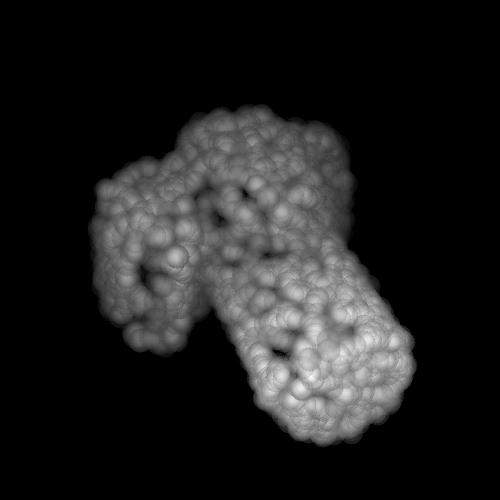} & \includegraphics[width=\widf]{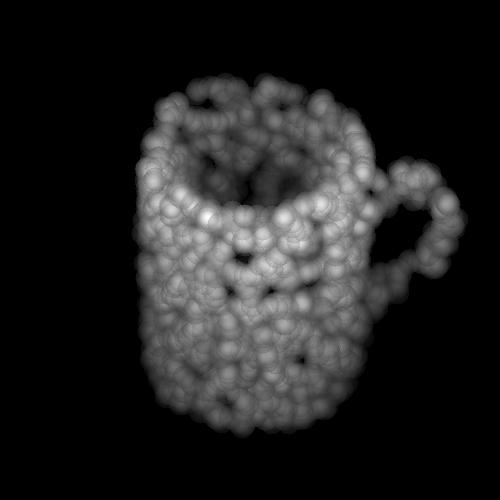} & \includegraphics[width=\widf]{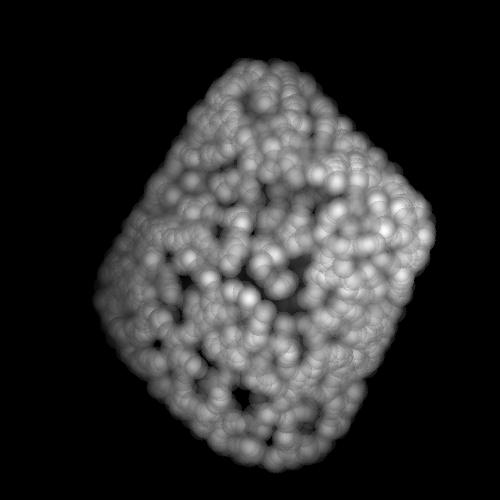} & \includegraphics[width=\widf]{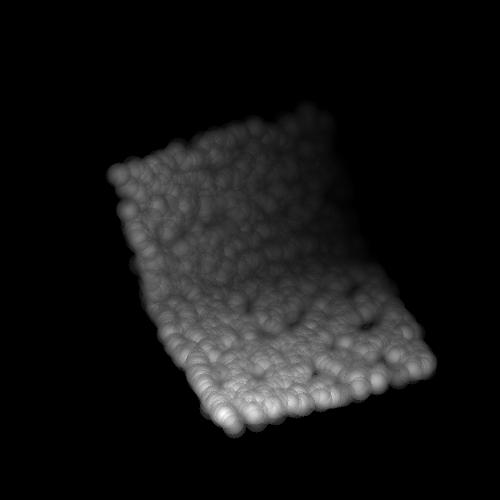} & \includegraphics[width=\widf]{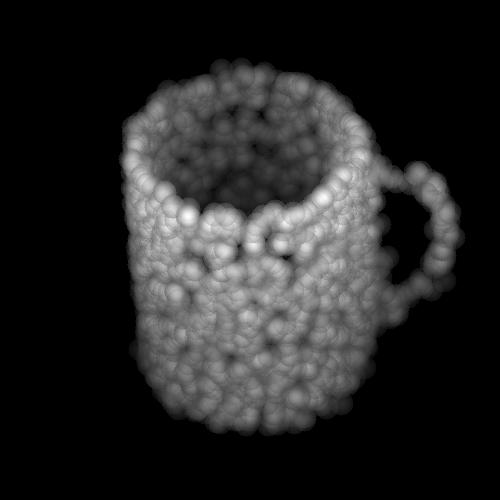} \\
\end{tabular}
}
\caption{Qualitative results on NOCS dataset. Top: input image. Middle: ground-truth 3D shape with ground-truth rotation. Bottom: predicted 3D shape with predicted rotation.}\label{fig:visNOCs}
\end{center}
%\vspace{-5px}
\end{figure*}

\noindent\textbf{Shape Reconstruction:} Table \ref{table_nocs_shape} shows that Glissando-Net with EMD significantly outperforms CPS in terms of mean Chamfer distance across all three categories. %, even though Glissando-Net is trained on all six categories. 
%Glissando-Net with Chamfer distance is worse than with EMD but still better than CPS. 
(CPS does not report shape reconstruction results on real data.) Even without the benefit of depth information, Glissando-Net outperforms CASS on three categories of the real data and achieves better overall performance. Glissando-Net with Chamfer distance is not as good as with EMD, but still outperforms the baselines. % as the EMD variant does. 
The poor performance of Glissando-Net$^*$ without the 3D encoder shows that the 3D encoder contributes substantially to shape reconstruction. %Particularly, Gilissando-Net without 3D encoder still performs better than CPS on  the synthetic data.
Table \ref{table_nocs_shape} shows that Glissando-Net both with Chamfer and EMD significantly outperforms OLD-Net\cite{fan2022object} in terms of mean Chamfer distance across all the categories on real and synthetic data except the laptop from the real data. OLD-Net\cite{fan2022object} does not report overall mean chamfer distance on the real data. NOCE \cite{lee2021category} only reports the overall Chamfer distance on the real data which is much worse than all other methods including Glissando-Net.
%When we train on both Synthetic and Real data, Glissando-Net performs even better.
%CASS in three categories and the overall performance on real data. %% PM moved above
% We 
% \begin{table}[thb]
% 	\caption{Shape Reconstruction Result on Synthetic Data from NOCS.}
% 	\label{table_nocs_synthetic_shape}
% 	\begin{center}
% 		\begin{tabular}{|c|c|c|c|}
% 			\hline
% 			\multirow{2} * {Methods} &
% 			\multicolumn{3}{|c|}{Mean Chamfer Distance in mm} \\
% 			\cline{2-4}
% 			& cup & bottle & camera \\ 
% 			\hline
% 			CPS \cite{manhardt2020cps} & 3.4 & 3.7 & 11.2 \\
% 			\hline
% 			Glissando-Net(EMD) & \textbf{0.39} & \textbf{0.63} & \textbf{1.16} \\
% 			\hline
% 			Glissando-Net(Chamfer) & 0.54 & 0.89 & 1.31 \\
% 			\hline
% 		\end{tabular}
% 	\end{center}
% \end{table}

%Besides CPS, we compare with another two baselines from CPS. CPS trains the network using L1 loss for rotation, translation and shape to create two baselines: %(1) set each weighting component as 1 (no weighting) (2) Multi-task weighting \cite{kendall2018multi}. 

\begin{figure*}[thb]
        \begin{center}
        \includegraphics[width=1.0\textwidth]{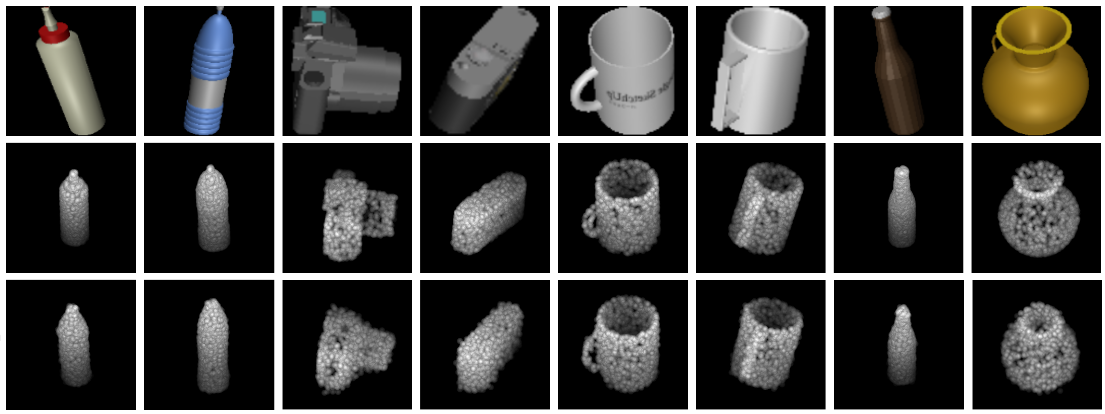} \\
    \end{center}
    %\vspace{-2pt}
    \caption{Qualitative results on \emph{synthetic} NOCS data. Top to bottom: input image, ground-truth 3D shape with ground-truth rotation, and our predicted 3D shape with predicted rotation.}\label{fig:vis_syn_more}
\end{figure*}

\begin{figure*}[thb]
    \begin{center}
        \includegraphics[width=1.0\textwidth]{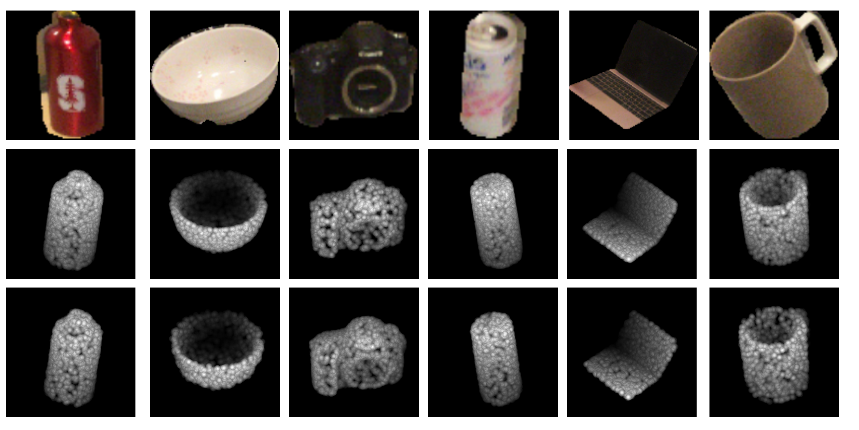} \\
    \end{center}
    %\vspace{-2pt}
    \caption{Qualitative results on \emph{real} NOCS data. Top to bottom: input image, ground-truth 3D shape with ground-truth rotation, and our predicted 3D shape with predicted rotation.} \label{fig:vis_real_more}
\end{figure*}

\begin{figure}[thb]
    \begin{center}
        \includegraphics[width=0.5\textwidth]{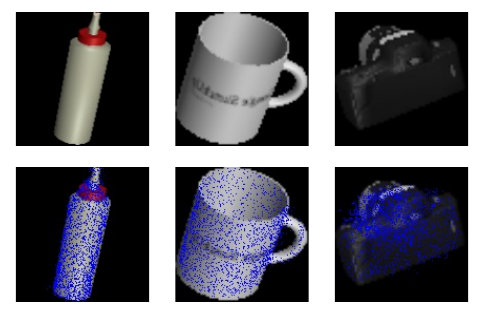} \\
    \end{center}
    %\vspace{-2pt}
    \caption{Overlay results on NOCS data. Top to bottom: input image, project the predicted point cloud to the input image.} \label{fig:overlay}
\end{figure}

\noindent\textbf{Pose Estimation:} The authors of CPS \cite{manhardt2020cps} implemented the NOCS network without the depth branch. % for a fair comparison. 
%But the pose estimation results of NOCS without depth is very bad. So we don't compare with NOCS without depth. 
We do not repeat these results here since they are not competitive. Table~\ref{table_nocs_pose} shows pose estimation results. 
On the synthetic subset, Glissando-Net with EMD performs better than CPS on all metrics in all categories, except for %the 10\degree\&10cm metric 
APP with $\alpha=0.5$ on the camera category. On the real subset, Glissando-Net with EMD performs better than CPS %on all metrics in all categories. When we train on both synthetic and real data, Glissando-Net performs the best on 
according to the 10\degree \& 10 cm metric in all categories and also better in most shape estimation metrics. Table \ref{table_nocs_pose_all} shows the overall pose estimation results across all of 6 categories. Glissando-Net ranks first on synthetic data and performs better than CPS but worse than CPS++ on real data. (CPS++ has been trained on large amounts of additional data, but in a self-supervised manner.)
Glissando-Net$^*$ without the 3D encoder performs worse than all other methods
%CPS, CPS++ and Glissando-Net 
on both synthetic and real data. 
OLD-Net \cite{fan2022object} and NOCE \cite{lee2021category} only report overall pose estimation results using the 10\degree \& 10 cm metric on the synthetic and real data. Glissando-Net performs much better than OLD-Net \cite{fan2022object} and NOCE \cite{lee2021category} on pose estimation.
%3D encoder is very important to pose estimation.
Figure \ref{fig:visNOCs} shows  qualitative results of predicted 3D shapes and poses. Figures~\ref{fig:vis_syn_more} and \ref{fig:vis_real_more} contain additional results on the synthetic and real subsets of the NOCS dataset \cite{wang2019normalized}. We project the predicted point cloud with an estimated pose to the input image. Figure \ref{fig:overlay} shows the overlayed visual results.

%Table \ref{table_nocs_real_pose} shows the pose estimation results on the real data. Glissando-Net with EMD performs the best on all metrics in all categories.

%\begin{table}[thb]
%	\caption{Pose Estimation Results on Real Data from NOCS}
%	\label{table_nocs_real_pose}
%	\begin{center}
%	    \scalebox{0.85}{
%		\begin{tabular}{|c|c|c|c|c|c|c|}
%			\hline
%			\multirow{2} * {Methods} &
%			\multicolumn{3}{|c|}{APP($\alpha$ = 0.2/0.5)} &
%			\multicolumn{3}{|c|}{10\degree \& 10 cm} \\
%			\cline{2-7}
%			&cup & bottle & camera & cup & bottle & camera \\ 
%			\hline
%			Multi-Task Weighting \cite{kendall2018multi} & 19.9\%/55.0\% & 0.0\%/2.1\% & 3.6\%/32.4\% & 0.7\% & 0.2\% & 0.0\% \\
%			\hline
%			CPS \cite{manhardt2020cps} & 36.8\%/64.6\% & 1.6\%/14.1\% & 19.6\%/56.7\% & 1.3\% & 9.5\% & 1.9\% \\
%			\hline
%			Glissando-Net(EMD) & \textbf{39.44\%/81.07\%} & \textbf{9.95\%/27.59\%} & \textbf{22.96\%}/52.34\% & \textbf{2.71\%} & \textbf{11.29\%} & \textbf{2.23\%} \\
%			\hline
%			Glissando-Net(Chamfer) & \textbf{38.84\%/68.03\%} & \textbf{3.44\%/15.25\%} & \textbf{19.82\%}/45.74\% & \textbf{2.64\%} & \textbf{7.04\%} & \textbf{1.49\%} \\
%			\hline
%		\end{tabular}
%		}
%	\end{center}
%\end{table}

\subsection{Pix3D Dataset Results} \label{subsec:experi_pix3d}
\begin{figure*}[tb]
\begin{center}
\scalebox{0.9}{
\begin{tabular}{ccccc}
\includegraphics[width=\widf]{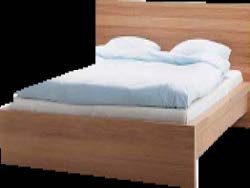} & \includegraphics[width=\widf]{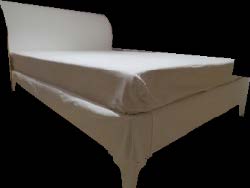} & \includegraphics[width=\widf]{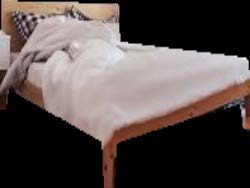} & \includegraphics[width=\widf]{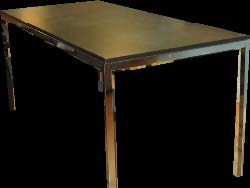} & \includegraphics[width=\widf]{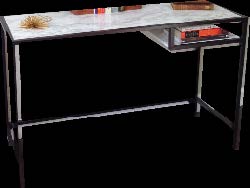} \\
\includegraphics[width=\widf]{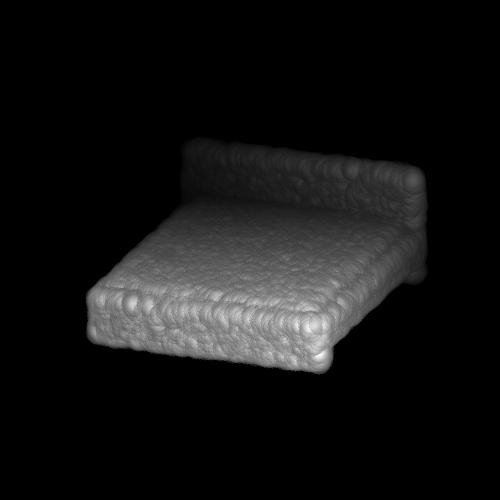} & \includegraphics[width=\widf]{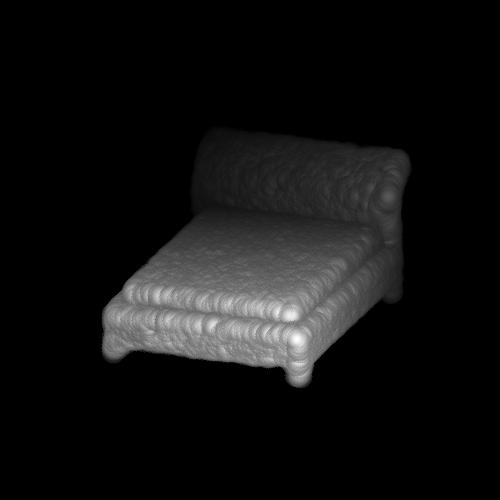} & \includegraphics[width=\widf]{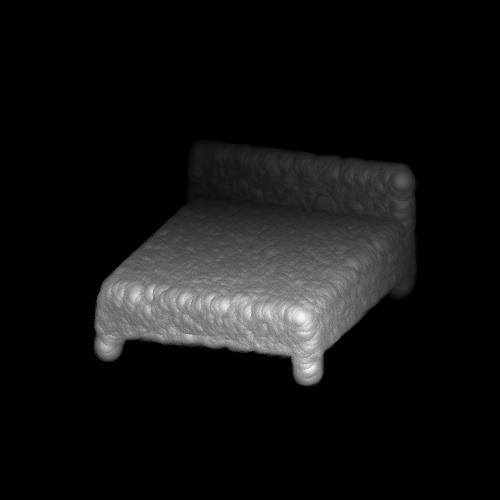} & \includegraphics[width=\widf]{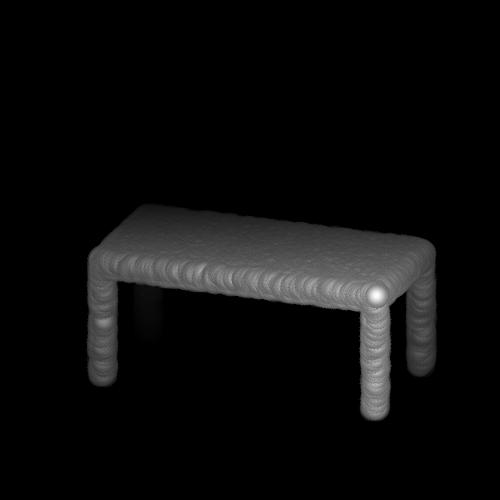} & \includegraphics[width=\widf]{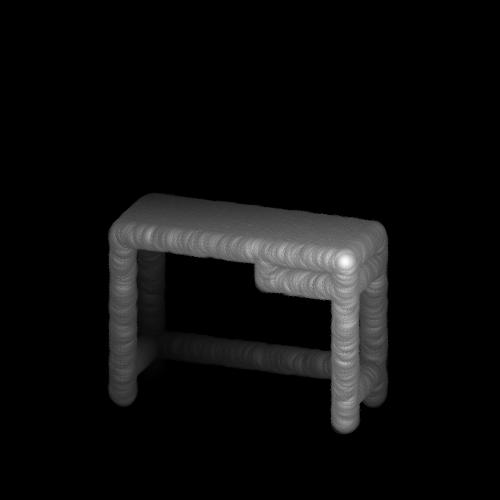} \\
\includegraphics[width=\widf]{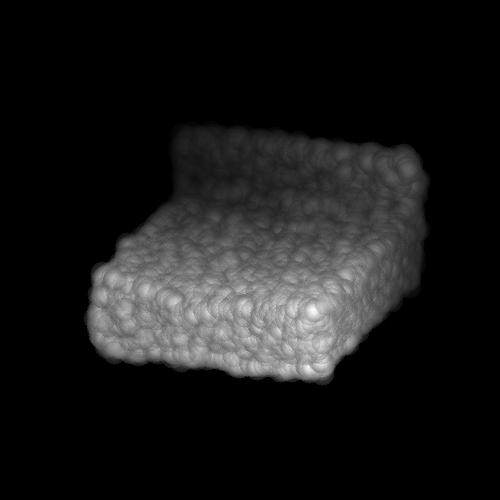} & \includegraphics[width=\widf]{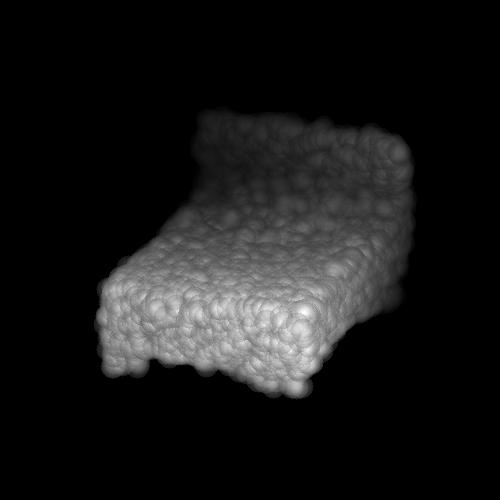} & \includegraphics[width=\widf]{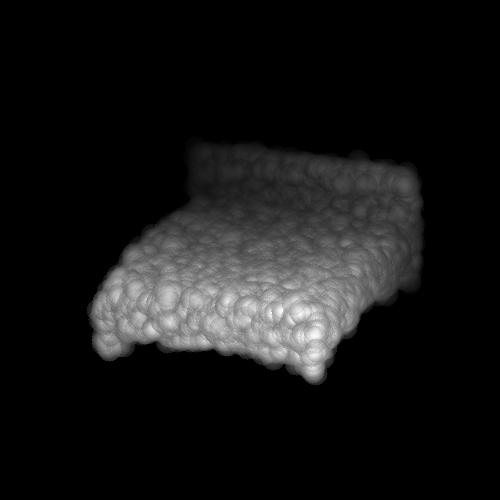} & \includegraphics[width=\widf]{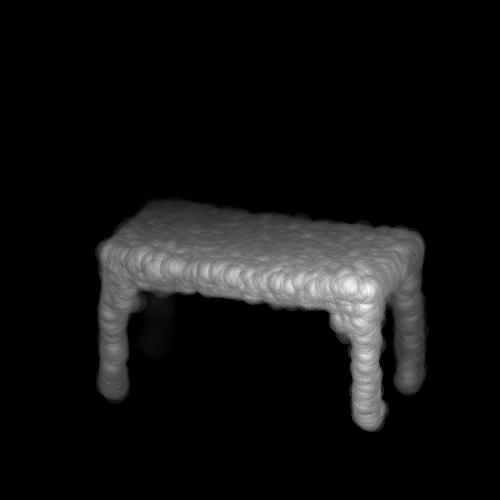} & \includegraphics[width=\widf]{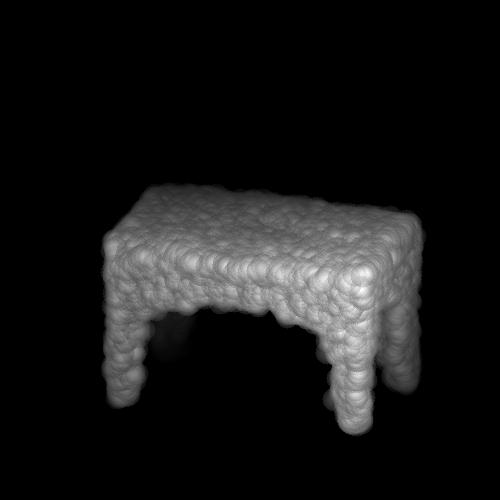} \\
\end{tabular}
}
\end{center}
\vspace{-5px}
\caption{Qualitative results on Pix3D: input (top row). ground-truth 3D shape (second row) and our predicted 3D shape (third row). The dataset does not provide pose estimation evaluation. Point clouds are shown in a pose chosen for visualization.}
\label{fig:visPix3d}
\end{figure*}

\begin{figure*}[thb]
      \begin{center}
        \includegraphics[width=0.9\textwidth]{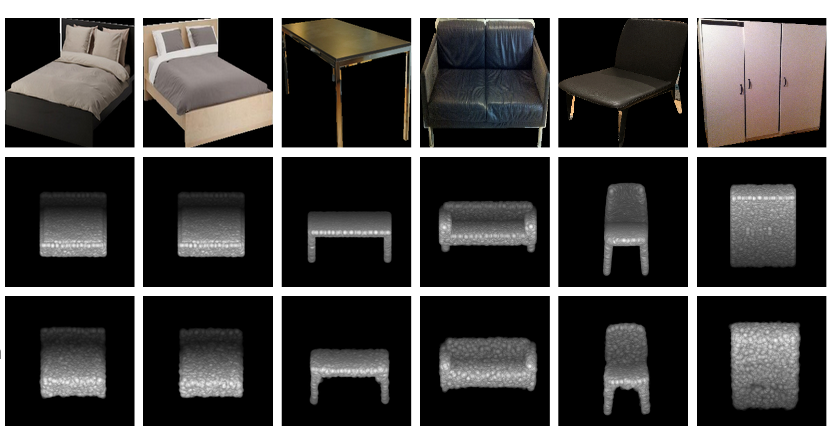} \\
    \end{center}
    %\vspace{-2pt}
    \caption{Qualitative results on Pix3D. Top to bottom: input image, ground-truth 3D shape with ground-truth rotation, and our predicted 3D shape with predicted rotation. \emph{Notice that many of the shapes are partially occluded in the images.}}  \label{fig:vis_pix3d_more}
\end{figure*}

We also evaluate our method on the Pix3D Dataset using the two train/test splits from \cite{Gkioxari_2019_ICCV}: $S_{1}$ and $S_{2}$ which are instance-level and category-level, respectively. We compare with \cite{nie2020total3dunderstanding} and two baselines from \cite{nie2020total3dunderstanding} for instance-level shape reconstruction, and with Mesh R-CNN \cite{Gkioxari_2019_ICCV} for category-level shape reconstruction. Since Mesh R-CNN does not report Chamfer distance results, we ran the authors' code and computed Chamfer distances on the generated meshes. (We exclude the tool and wardrobe categories due to our inability to reproduce reasonable results.) Table \ref{table_pix3d_instance_shape} summarizes the results. For instance-level shape reconstruction, Glissando-Net achieves the best results on all categories, except for the tool for which there are insufficient samples.
%from four to five different kinds of tools. 
For category-level shape reconstruction, Glissando-Net performs better than Mesh R-CNN. Compared with instance-level methods, Glissando-Net on $S_{2}$ performs a little worse than Total3D \cite{nie2020total3dunderstanding}, but still better than \cite{groueix2018atlasnet} and \cite{pan2019deep}, on $S_{1}$. The mean error of Glissando-Net on $S_{2}$ is computed over 7 categories. The mean error over 9 categories is 14.39 which is still lower than the mean error of Mesh R-CNN on $S_{2}$ over 7 categories.
Figures~\ref{fig:visPix3d} and \ref{fig:vis_pix3d_more} show several qualitative shape reconstruction results from the Pix3D dataset \cite{sun2018pix3d}.

\subsection{Ablation Studies}
We perform three ablation studies on a subset of the NOCS synthetic dataset. Specifically, for the first two ablation studies, we randomly sample 10K images and their point clouds from the synthetic training dataset for training, and 200 images from each of three categories 
from the testing dataset for validation. For the last ablation study, we train and test the model on synthetic data from the NOCS dataset.

\noindent\textbf{Length of the latent code $\mathbf{z}$ in the VAE}: We evaluate the effect of the latent code $\mathbf{z}$ with different vector lengths: 32, 64, 128, 256, 512, 1024. Table \ref{table_nocs_ab_code2} shows the results of shape reconstruction and pose estimation with different sizes of the latent code. As expected, larger codes yield better results. When the code size is larger than 256, the performance does not change a lot. For computational efficiency, we set the code size to 256 for all other experiments in the paper. 

\begin{table}[h!]
    \caption{Ablation study on code size}\label{table_nocs_ab_code2}
	\begin{center}
	    %\scalebox{0.9}{
		\begin{tabular}{|c|c|c|c|c|}
			\hline
			\tabincell{c}{Code \\ Size} & \tabincell{c}{Shape \\ Chamfer mm $\downarrow$} & \tabincell{c}{APP \\ ($\alpha$ = 0.5) $\uparrow$} & \tabincell{c}{APP \\ ($\alpha$ = 0.2) $\uparrow$} & \tabincell{c}{10\degree \\ 10 cm $\uparrow$} \\ 
			\hline
			32 & 0.98 & 52.50\% & 28.50\% & 6.67\% \\
			\hline
			64 & 0.90 & 57.67\% & 30.83\% & 14.50\% \\
			\hline
			128 & 0.87 & 57.33\% & 31.17\% & 16.83\% \\
			\hline
			256 & 0.81 & 58.50\% & 35.50\% & 18.50\% \\
			\hline
			512 & 0.81 & 58.83\% & 34.50\% & 17.91\% \\
			\hline
			1024 & 0.80 & 58.67\% & 35.70\% & 18.84\% \\
			\hline
		\end{tabular}
		%}
	\end{center}
	%\vspace{-8pt}
\end{table}

\noindent\textbf{Effects of the Pose Sub-network}: We experimented with different positions of the pose regression network: after the shape decoder or after the image encoder. We also trained our network without the pose network. Table \ref{table_nocs_ab_pose} shows the results. From the top two rows, we can see that shape reconstruction can help pose estimation. The results in the second and third columns indicate that jointly estimating pose can help shape reconstruction. We also tried to use a single network to predict both rotation and translation. This single network has five fully connected layers with the last layer outputting a 4D quaternion and 3D translation. The last row of Table \ref{table_nocs_ab_pose} shows the results. Pose estimation is worse compared to using separate networks for rotation and translation estimation.%It makes the pose estimation accuracy worse to use only one network instead of two networks.

\begin{table}[thb]
    \caption{Ablation study: with/without pose estimation} \label{table_nocs_ab_pose}
	\begin{center}
	    \scalebox{0.9}{
		\begin{tabular}{|c|c|c|c|c|}
			\hline
			 & \tabincell{c}{Shape \\ Chamfer mm $\downarrow$} & \tabincell{c}{APP \\ ($\alpha$ = 0.5) $\uparrow$} & \tabincell{c}{APP \\ ($\alpha$ = 0.2) $\uparrow$} & \tabincell{c}{10\degree \\ 10 cm $\uparrow$} \\ 
			\hline
		    \tabincell{c}{Pose after \\ shape decoder} & 0.81 & 58.50\% & 35.50\% & 18.50\% \\
			\hline
			\tabincell{c}{Pose after \\ image encoder} & 0.81 & 53.50\% & 30.60\% & 15.67\% \\
			\hline
			Without pose & 0.83 & - & - & - \\
			\hline
            Pose in single network & 0.80 & 49.76\% & 25.34\% & 10.20\% \\
            \hline
		\end{tabular}
		}
	\end{center}
	%\vspace{-8pt}
\end{table}

\noindent\textbf{Effects of the Transform Modules}: We performed an additional ablation study on our feature transforms. We made Glissando-Net$_{DisEn}$ by disconnecting the feature transform module between RGB decoder and point cloud encoder, and Glissando-Net$_{DisDe}$ by disconnecting the feature transform module between RGB decoder and point cloud decoder (see Fig. \ref{fig:workflow}). We trained all variants on the NOCS dataset using the EMD distance in the loss. Glisando-Net$^*$ is the baseline without the 3D encoder as in Tables 1-3. As shown in Table \ref{table_nocs_ab_tm}, the second feature transform module is more critical than the first one and the 3D encoder. This table is an extension of Tables1 \& 3 and contains shape and pose estimation results. (The averages are over all samples in the \emph{six} NOCS categories, covering more synthetic data than Table 1.)

\begin{table}[thb]
    \caption{Ablation study for feature transform module.} \label{table_nocs_ab_tm}
	\begin{center}
	    \scalebox{0.9}{
    	\begin{tabular}{|c|c|c|c|c|}
			\hline
			\multirow{2} * {Methods} &
			\multicolumn{2}{|c|}{Synthetic Data} & \multicolumn{2}{|c|}{Real Data} \\
			\cline{2-5}
			%& 10$^{o}$ %\degree 
			%\& 10 cm $\uparrow$ & APP ($\alpha$ = 0.2/0.5) $\uparrow$ & 10$^{o}$ \& 10 cm $\uparrow$ & APP ($\alpha$ = 0.2/0.5) $\uparrow$ \\ 
			& Chamfer & 10$^{o}$ \& 10 cm & Chamfer & 10$^{o}$ \& 10 cm \\
			\hline
			Glissando-Net$^*$ & 3.14 & 21.21\% & 2.85 & 2.97\% \\
			\hline
			Glissando-Net & 0.62 & \textbf{31.86\%} & 0.71 & 19.08\% \\
			\hline
			Glissando-Net $_{DisEn}$ & 0.94 & 24.98\% & 1.24 & 4.16\% \\
			\hline
			Glissando-Net $_{DisDe}$ & 3.44 & 2.06\% & 6.16 & 1.60\% \\
			\hline
		\end{tabular}
		}
	\end{center}
% 	\vspace{-8pt}
% 	%Glissando-Net (EMD)$^*$ represents Glissando-Net (EMD) without the 3D encoder.}\label{table_nocs_pose_all}
% 	\vspace{-15pt}
\end{table}

%\begin{figure}[thb]
    %\centering
    %\includegraphics[width=1.0\linewidth]{figure/VisOcc.pdf}
    %\vspace{-16pt}
    %\caption{Reconstructed 3D shape with occluded RGB input. Top, left to right:
    %we show the 3D shape with 
    %original RGB image and the ones occluded from bottom, center, left, right, and top. Bottom: Corresponding reconstructed shapes, shown in the same pose for visualization.}
    %\label{fig:visOcc}
    %\vspace{-12pt}
%\end{figure}

%\newcommand\widf{.40\columnwidth}
\begin{figure}[tb]
\begin{center}
\scalebox{0.5}{
\begin{tabular}{cccc}
\includegraphics[width=\widf]{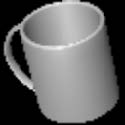} & \includegraphics[width=\widf]{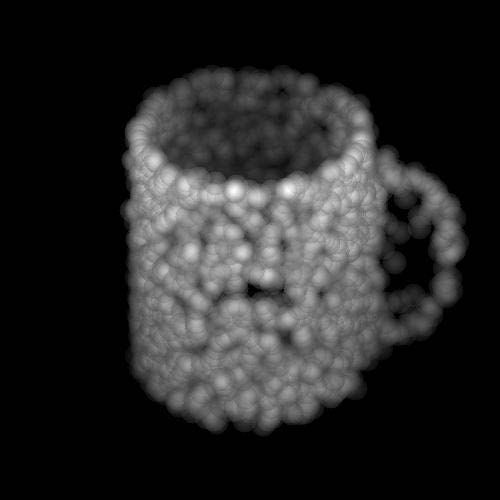} & \includegraphics[width=\widf]{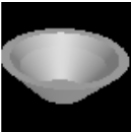} & \includegraphics[width=\widf]{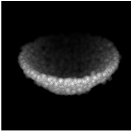} \\ 
\includegraphics[width=\widf]{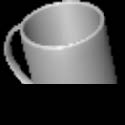} & \includegraphics[width=\widf]{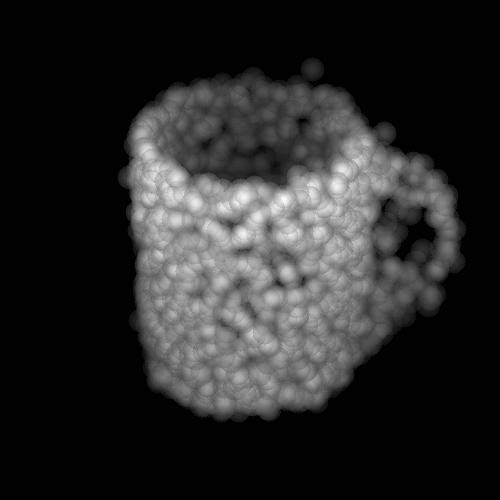} & \includegraphics[width=\widf]{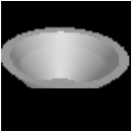} & \includegraphics[width=\widf]{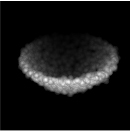} \\
\includegraphics[width=\widf]{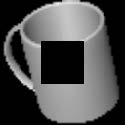} & \includegraphics[width=\widf]{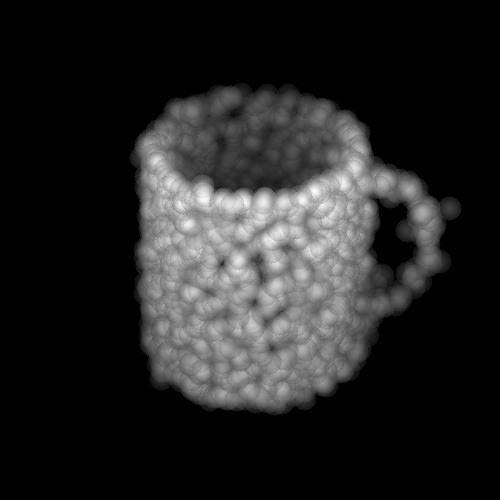} & \includegraphics[width=\widf]{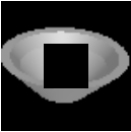} & \includegraphics[width=\widf]{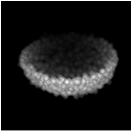} \\
\includegraphics[width=\widf]{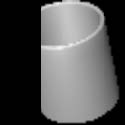} & \includegraphics[width=\widf]{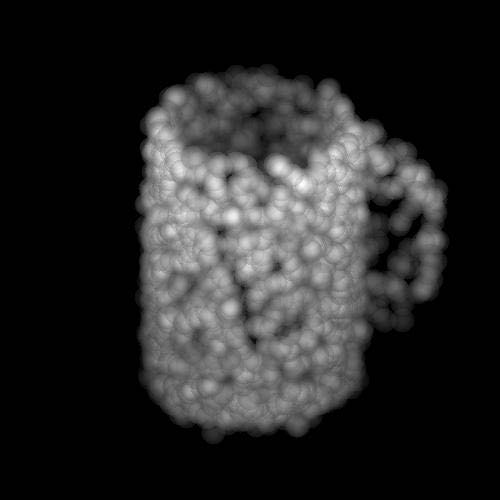} & \includegraphics[width=\widf]{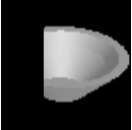} & \includegraphics[width=\widf]{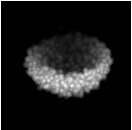} \\
\includegraphics[width=\widf]{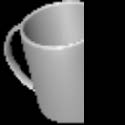} & \includegraphics[width=\widf]{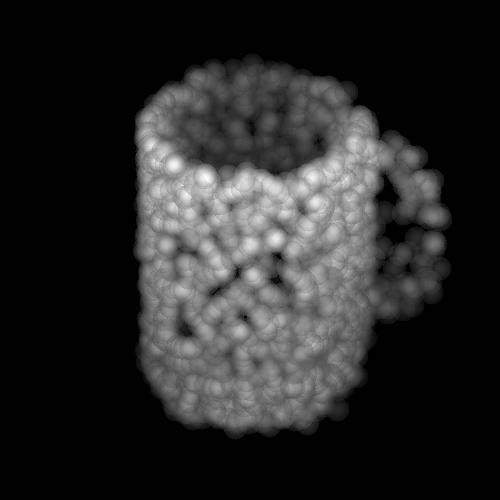} & \includegraphics[width=\widf]{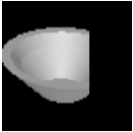} & \includegraphics[width=\widf]{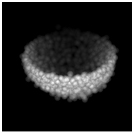} \\
\includegraphics[width=\widf]{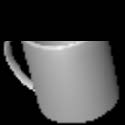} & \includegraphics[width=\widf]{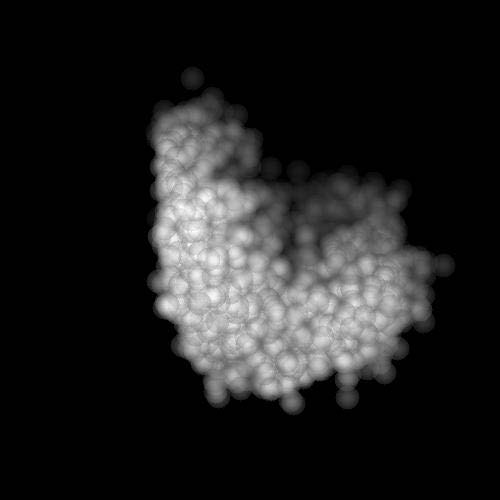} & \includegraphics[width=\widf]{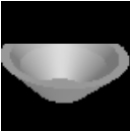} & \includegraphics[width=\widf]{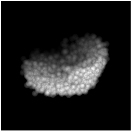} \\
\end{tabular}
}
\end{center}
\vspace{-5px}
\caption{Reconstructed 3D shape with occluded RGB input. Column 1 and 3, top to bottom: original RGB image and the ones occluded from bottom, center, left, right, and top. Column 2 and 4: Corresponding reconstructed shapes, shown in the same pose for visualization.}
\label{fig:visOcc}
\end{figure}

%\begin{figure}[thb]
    %\centering
    %\includegraphics[width=0.8\linewidth]{figure/VisPix3dNew.jpg}
    %\caption{Qualitative results on Pix3D: the ground-truth 3D shape (second row) and our predicted 3D shape (third row). The dataset does not provide pose estimation evaluation. Point clouds are shown in a pose chosen for visualization.}
    %\label{fig:visPix3d}
    %\vspace{-12pt}
%\end{figure}

%\newcommand\widf{.40\columnwidth}

\begin{figure*}[thb]
        \begin{center}
        \includegraphics[width=1.0\textwidth]{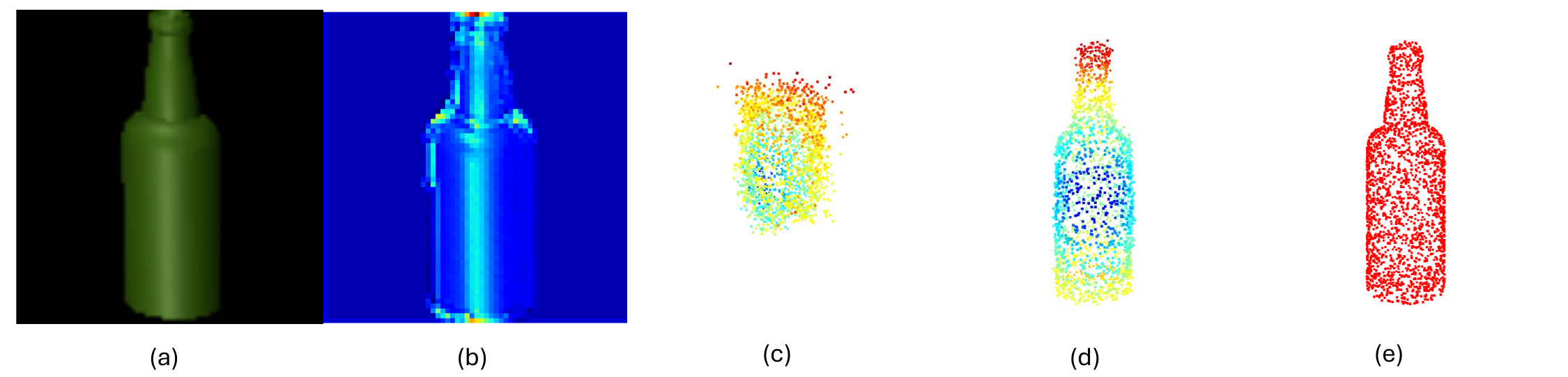} \\
    \end{center}
    \vspace{-5pt}
    \caption{Feature Map Visualization for 2D and 3D. (a) shows the input RGB image. (b) shows the 2D image feature map. (c) shows the point cloud generated without the RGB encoder and decoder. (d) shows the point cloud generated with the RGB encoder and decoder. (e) shows the ground truth 3D shape.}\label{fig:feature_map_2d3d}
\end{figure*}

\subsection{Feature Map Visualization}
We visualize the RGB features and the point clouds generated with and without RGB input in Fig. \ref{fig:feature_map_2d3d}. Fig. \ref{fig:feature_map_2d3d}(a) shows the input RGB image. Fig. \ref{fig:feature_map_2d3d}(b) illustrates the 2D image feature map, which was obtained from the penultimate layer of the RGB decoder. Fig. \ref{fig:feature_map_2d3d}(c) presents the point cloud generated without the RGB encoder, RGB decoder, or the feature transform module, relying solely on the point cloud encoder and decoder. In contrast, Fig. \ref{fig:feature_map_2d3d}(d) displays the point cloud generated when the RGB encoder, RGB decoder, and the feature transform module are included. Finally, Fig. \ref{fig:feature_map_2d3d}(e) shows the ground truth 3D shape. To better highlight the role of the RGB encoder, RGB decoder, and the feature transform module, we selected the topmost 2D pixel of the bottle in the input image (Fig. \ref{fig:feature_map_2d3d}(a)). This selected pixel was then projected into the corresponding point cloud. Subsequently, the distances between this point feature and the features of all other points in the point cloud were calculated to generate a heatmap, which was visualized on the point cloud in Fig. \ref{fig:feature_map_2d3d}(c) and Fig. \ref{fig:feature_map_2d3d}(d). To illustrate the importance of the RGB encoder, RGB decoder, and the feature transform module, we conducted the visualization using an image from the training set rather than the testing set. The reason for this choice is that, during the testing phase, the point cloud encoder is absent, leaving only the RGB encoder. Consequently, it would not be feasible to remove the RGB input and assess the contribution of 2D image information to the overall performance. The results demonstrate that the point cloud shown in Fig. \ref{fig:feature_map_2d3d}(c) contains only the lower part of the bottle, and the overall shape is inaccurate. In contrast, the point cloud shown in Fig. \ref{fig:feature_map_2d3d}(d), which benefits from the inclusion of RGB input, more closely resembles the ground truth 3D shape illustrated in Fig. \ref{fig:feature_map_2d3d}(e).

\subsection{Results on Partially Occluded Objects}
Glissando-Net can handle partially occluded objects thanks to the latent shape space in addition to the RGB appearance network. We randomly pick 2K images from the NOCS synthetic testing data and render five kinds of occlusion (top, bottom, left, right and center) on each image, using a black block with 1/3 width or height of the original image. We tested these 10K partially occluded images as well as the corresponding 2K occlusion-free original images using the network trained on the synthetic training data of NOCS \emph{without occlusion}. Some of the 3D shape reconstruction results are shown in Fig.~\ref{fig:visOcc}. Although the occluded shapes are not as precise as their occlusion-free counterparts, the network still completes the shape for the unseen regions exceptionally. We believe that this is because our VAE successfully learns categorical shape space as a strong prior to compensating for the incomplete appearance information. 
We speculate that occluding the top part of a cup is more challenging since the top is harder to predict from the visible parts of the shape.
The average Chamfer distance from ground truth is 0.56 mm on images without occlusion and 1.17 mm on images with occlusion. The average 10\degree 10 cm accuracy is 34.26\% on images without occlusion and 20.54\% on images with occlusion.

We also present quantitative results on shape reconstruction and pose estimation from occluded RGB images. We report shape reconstruction and pose estimation results separately for the five directions of occlusion in Table~ \ref{table_statistics_occlude}. The quantitative results are consistent with the visualization results in Figure~\ref{fig:visOcc}. When the top of the object is occluded, Glissando-Net cannot reconstruct its 3D shape correctly and cannot estimate its pose accurately.

\begin{figure*}[thb]
        \begin{center}
        \includegraphics[width=1.0\textwidth]{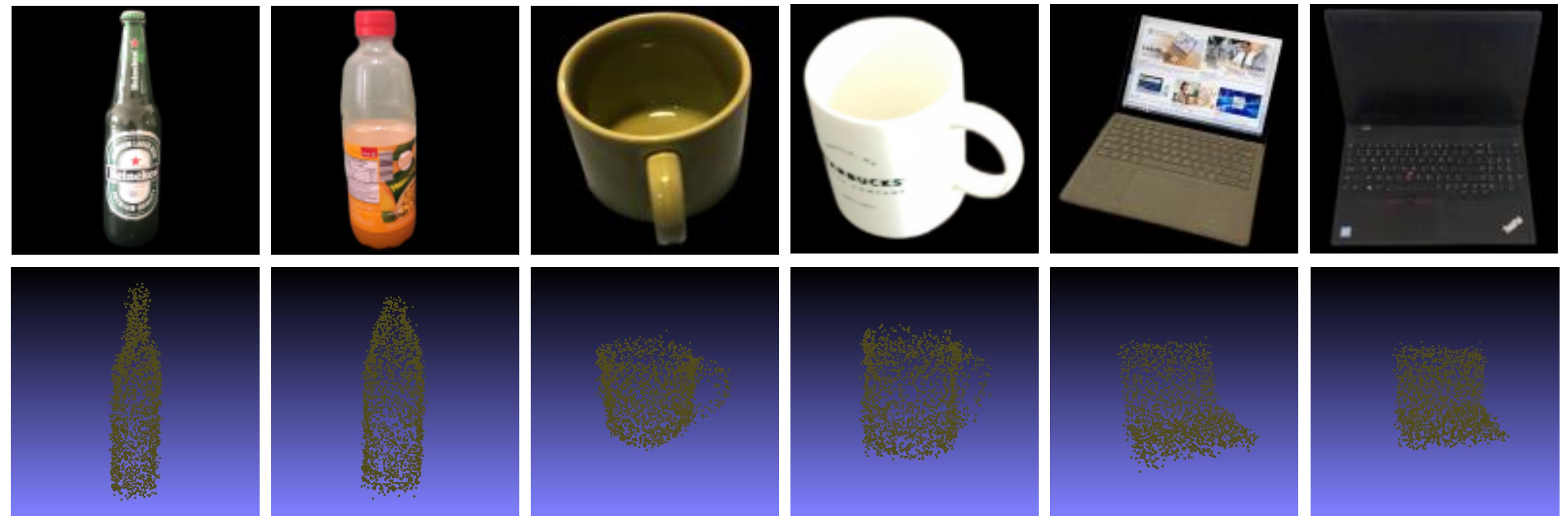} \\
    \end{center}
    \vspace{-5pt}
    \caption{Qualitative Results on Objectron Dataset. The first row shows the input image. The second row shows the reconstructed 3D shape.}\label{fig:objectron}
\end{figure*}

\subsection{Qualitative Results on Objectron Dataset}
To evaluate the cross-domain generalization ability of our model, we conduct zero-shot testing by directly applying the model trained on the NOCS dataset \cite{wang2019normalized} to the Objectron dataset \cite{objectron2021} without any retraining. We select three object categories from the Objectron dataset \cite{objectron2021}: bottle, cup, and laptop. We extract video frame images using the official code provided on the Objectron GitHub repository. Since our model requires input images with foreground objects masked, and Objectron does not provide ground truth masks, we use Rembg \cite{qin2020u2} to generate the foreground masks for each frame. We then use the bounding box of the foreground to crop the region of interest and resize it to 128x128, matching the input resolution required by our model. The processed images are fed into our model, which is pre-trained on the NOCS dataset \cite{wang2019normalized}, for testing. The point clouds provided by the Objectron dataset \cite{objectron2021} are extremely sparse, containing only 200 to 400 points per object, and these points are not uniformly distributed. Due to this sparsity, it is not feasible to obtain ground truth point clouds, and consequently, we are unable to obtain reasonable quantitative results. However, qualitative results of 3D shape reconstruction are shown in Figure \ref{fig:objectron}. These results demonstrate that, despite not being trained on Objectron, our model achieves reasonable 3D shape reconstruction. This highlights the robustness of our model across different domains.

\begin{table}[h!]
    \caption{Shape reconstruction and pose estimation results on occluded RGB inputs from the NOCS synthetic dataset. The table shows the mean Chamfer distance in mm for each occlusion direction and the average over all direction in the last column. As a reminder, the mean Chamfer distance without occlusion is 0.56 mm and the 10\degree 10 cm accuracy without occlusion is 34.26\%.} \label{table_statistics_occlude}
	\begin{center}
	    \scalebox{0.8}{
		\begin{tabular}{|c|c|c|c|c|c|c|}
			\hline
			 & bottom & center & left & right & top & overall \\ 
			\hline
			Chamfer Distance(mm) $\downarrow$ & 0.85 & 0.74 & 1.13 & 0.89 & 2.23 & 1.17 \\
            \hline
            \tabincell{c}{10\degree 10 cm $\uparrow$} & 22.36\% & 26.77\% & 21.96\% & 22.07\% & 9.56\% & 20.54\% \\
			\hline
		\end{tabular}
		}
	\end{center}
\end{table}

%\LGcomments{LG: last column occlusion result should we just not show it? or explain a little bit about it?}

%\begin{table}[thb]
%	\caption{3D shape reconstruction results on images with/without occlusion.}
%	\label{table_occlusion}
%	\begin{center}
%		\begin{tabular}{|c|c|}
%			\hline
%			 & Shape(Chamfer mm) \\ 
%			\hline
%			images with occlusion & 1.17 \\
%			\hline
%			images without occlusion & 0.56 \\
%			\hline
%		\end{tabular}
%	\end{center}
%\end{table}

%------------------------------------------------------------------------
\subsection{Limitations}

\begin{figure}[thb]
    \begin{center}
        \includegraphics[width=0.4\textwidth]{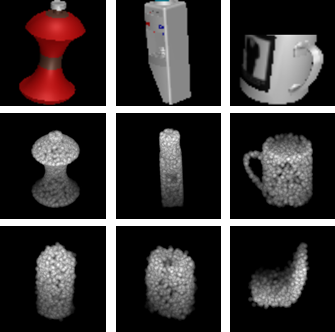}
    \end{center}
    %\vspace{-2pt}
    \caption{Some examples where our algorithm fails. The two examples on the left are form the bottle category, while the example on the right is from the mug category with occlusion from the top. The left two examples are of very unusual shapes. The rightmost example is occluded at the top. }\label{fig:vis_fail_more}
\end{figure}

Figure \ref{fig:vis_fail_more} shows examples on which our method fails. As mentioned in the conclusions, while Glissando-Net has proven to be robust to occlusion in general, it is less robust when the top of the object is occluded. We hypothesize that this is because the network cannot benefit from symmetry or closure in these cases. It seems easier to compensate for occlusion on the left, right or bottom. Occlusions from the top are particularly challenging due to the significant loss of critical shape information, such as the mug's opening and handle, which are often partially or entirely obscured. In contrast, side occlusions tend to obscure less critical features, allowing for better shape inference based on visible parts. To improve handling of top occlusions, enhancing the network's ability to infer missing parts from partial views or integrating additional data, such as multiple viewpoints or temporal sequences, could be beneficial. This approach would provide a more comprehensive understanding of the object's structure, allowing the network to make more informed predictions even with significant occlusions. Occluding the middle of the object is not catastrophic since the network is able to complete the shape. The two leftmost examples in Figure~\ref{fig:vis_fail_more} demonstrate that input images depicting very unusual shapes cause difficulties since Glissando-Net has not observed any similar examples in training. %Compare the bottles in Figure~\ref{fig:vis_fail_more} with those in Figures~1 and 3, and Figures \ref{fig:vis_syn_more} and \ref{fig:vis_real_more}.

%\begin{figure*}[thb]
        %\begin{center}
        %\includegraphics[width=0.8\textwidth]{figure/VisOcc_more.png}
    %\end{center}
    %\vspace{-2pt}
    %\caption{Reconstructed 3D shapes from occluded RGB inputs. From left to right, we show the reconstructed 3D shape from:  the original RGB image and the ones occluded at the bottom, center, left, right, and top. Shapes are aligned in the same pose for visualization.}\label{fig:vis_occ_more}
%\end{figure*}

%------------------------------------------------------------------------

%------------------------------------------------------------------------
\section{Conclusions} \label{sec:conclusions}
We have presented a new deep learning model, Glissando-Net, for simultaneously estimating 6D pose and reconstructing 3D shape at the category level from a single RGB image. This is a research topic largely unexplored in the literature. The only work that attempts to achieve both goals from \emph{real} RGB inputs, to the best of our knowledge, is CPS~\cite{manhardt2020cps,manhardt2020cpsv3}. Two key design differences %in the deep networks 
enabled Glissando-Net to obtain more accurate prediction of both pose and shape.

First, CPS directly regresses object pose from the image. As a result the predicted pose cannot leverage any prior information conveyed by the point clouds that are only available during training. In contrast, Glissando-Net predicts both pose and shape in the decoder stage of the point cloud VAE, enabling pose prediction to better exploit priors %information 
learned from point clouds. Second, to more effectively utilize features from the RGB image, instead of just concatenating the codes, the feature maps extracted from the image encoder are also concatenated to the feature maps in the decoder of the point cloud VAE. This way, the image features are more thoroughly leveraged to achieve more accurate estimation.
As validated by our experiments, these contributed to higher prediction accuracy of both object poses and 3D shapes by Glissando-Net, when compared with CPS. Our experiments also demonstrate that the proposed method is relatively robust to occlusion.
{
\bibliographystyle{IEEEtran}
%\bibliography{IEEEfull}

}


%\end{IEEEbiography}


%\newpage


\begin{IEEEbiography}
[{\includegraphics[width=1in,height=1.25in,clip,keepaspectratio]{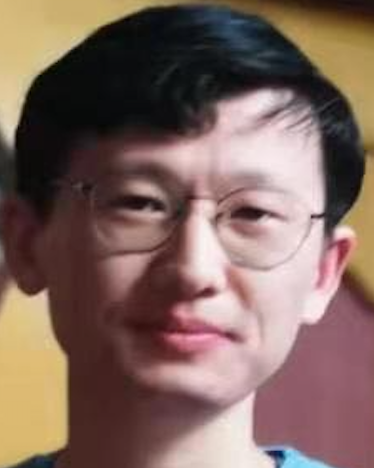}}]{Bo Sun}
is a researcher at Adobe Inc. He earned his PhD from the Stevens Institute of Technology and earned his Bachelor and Master degree from Nankai University, China. His research interests span 3D reconstruction from images, robotics and panoptic segmentation.
\end{IEEEbiography}

\begin{IEEEbiography}
[{\includegraphics[width=1.0in,height=1.25in,clip,keepaspectratio]{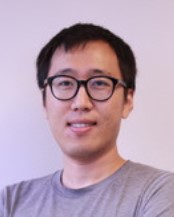}}]{Hao Kang} is a Senior Research Scientist at ByteDance Inc., specializing in Generative AI. Prior to this, he served as a Senior Researcher at Wormpex AI Research from 2019 to 2024. He holds a Ph.D. from Purdue University, a Master's degree from Stevens Institute of Technology, and a Bachelor's degree from Beijing Jiaotong University in China. His research interests encompass various aspects of visual computing, including generative AI, computer graphics, computer vision, and human-computer interaction.  
\end{IEEEbiography}

\begin{IEEEbiography}
[{\includegraphics[width=1.0in,height=1.25in,clip,keepaspectratio]{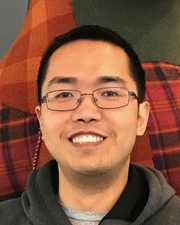}}]{Li Guan}
received his Ph.D. in Computer Science at the University of North Carolina at Chapel Hill in 2009. He is currently a Research Manager at Meta Reality Labs Research on sensor signal modeling and fusion to hand/glove poses and interaction in VR/AR. This work was done while he was a principal research manager at Wormpex AI Research LLC, leading projects to develop 3D solutions for store environment modeling to improve the efficiency of store operations. He is the author of over 20 refereed papers in 3D computer vision papers and more than 17 patents. He is in the organizing committee of the OmniCV (360/ominidirectional computer vision) workshop affiliated with the IEEE Conference on Computer Vision and Pattern Recognition (CVPR) 2020-2024.
\end{IEEEbiography}

\begin{IEEEbiography}
[{\includegraphics[width=1.0in,height=1.25in,clip,keepaspectratio]{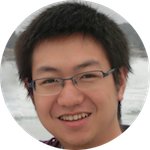}}]{Haoxiang Li}
received the BS degree in computer science from the University of Science and Technology of China, in 2010. He received the PhD degree from the Department of Computer Science at Stevens Institute of Technology, in 2016. He is currently Chief Scientist with Pixocial Technology. Before that, he was director of research with Wormpex AI Research, a Researcher with AiBee and Adobe Research. His research interests include human face processing and analysis, computer vision, and robotics. He was an area chair for IEEE Winter Conference on Applications of Computer Vision (WACV) 2020 and 2022.
\end{IEEEbiography}

\begin{IEEEbiography}
[{\includegraphics[width=1in,height=1.25in,clip,keepaspectratio]{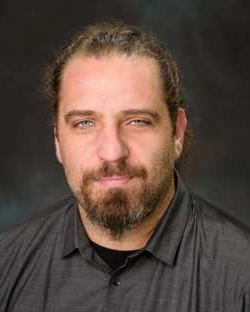}}]{Philippos Mordohai}
is a professor of Computer Science at Stevens Institute of Technology. He earned his PhD from the University of Southern California and held postdoctoral appointments at the University of North Carolina and the University of Pennsylvania. His research interests span 3D reconstruction from images and video, 3D segmentation and recognition, perception for robotics, and active vision. He serves as an associate editor for the IEEE Transactions on Pattern Analysis and Machine Intelligence, while he held similar roles for the Image and Vision Computing journal and Computer Vision and Image Understanding in the past. He has served as area chair for CVPR, ICCV and ECCV multiple times and program co-chair of the International Conference on 3D Vision (3DV) in 2019. Dr. Mordohai will be one of the program chairs of ICCV 2027.
\end{IEEEbiography}

\begin{IEEEbiography}
[{\includegraphics[width=1.0in,height=1.25in,clip,keepaspectratio]{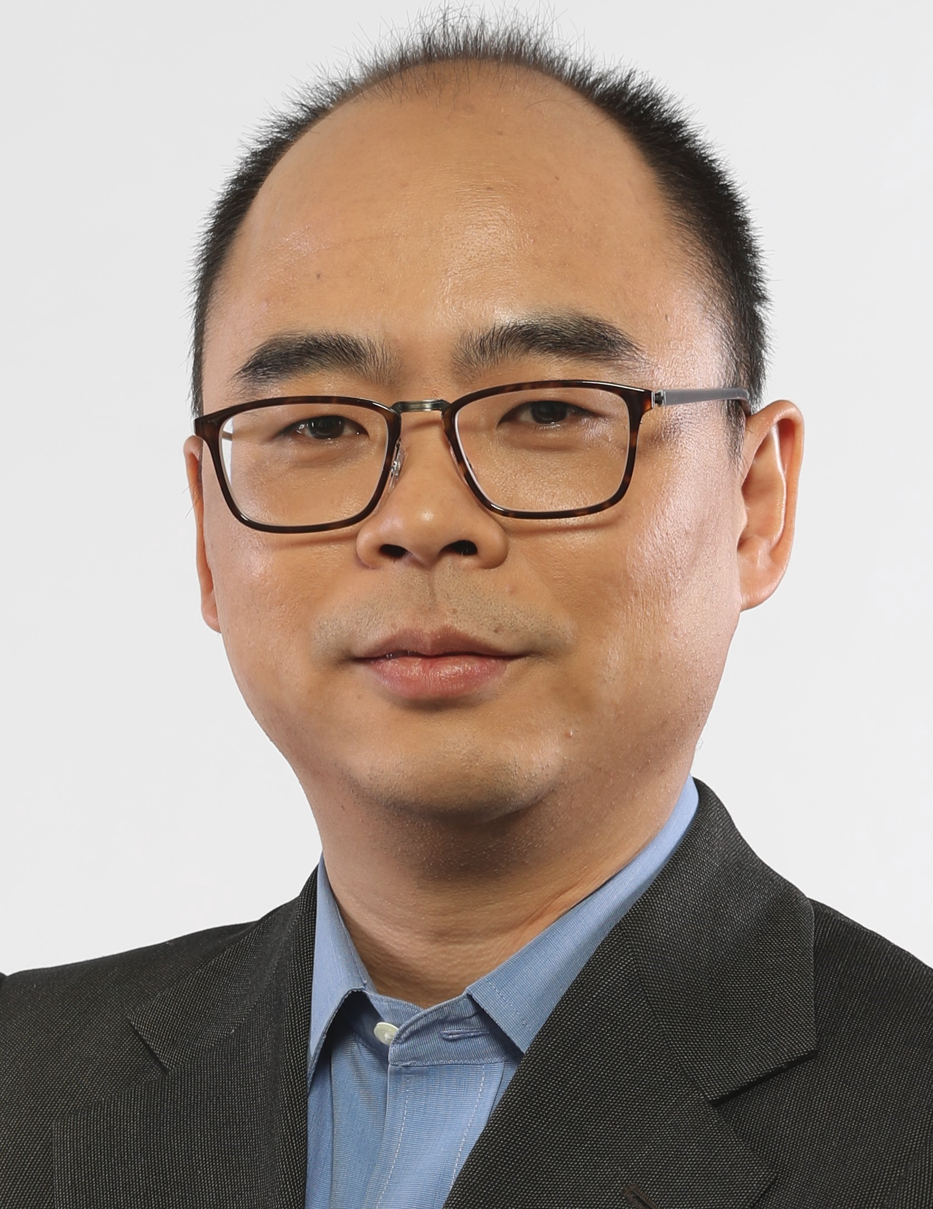}}]{Gang Hua} (M’03-SM’11-F’19) was enrolled in the Special Class for the Gifted Young of Xi’an Jiaotong University (XJTU), Xi’an, China, in 1994 and received the B.S. degree in Automatic Control Engineering from XJTU in 1999. He received the M.S. degree in Control Science and Engineering in 2002 from XJTU, and the Ph.D. degree in Electrical Engineering and Computer Science at Northwestern University, Evanston, Illinois, USA, in 2006. He is currently the Vice President of the Multimodal Experiences Research Lab with Dolby Laboratories. He was the Vice President and Chief Scientist of Wormpex AI Research, the US research branch of Convenience Bee. Before that, he served in various roles at Microsoft (2015-18) as the Science/Technical Adviser to the CVP of the Computer Vision Group, Director of Computer Vision Science Team in Redmond and Taipei ATL, and Principal Researcher/Research Manager at Microsoft Research. He was an Associate Professor at Stevens Institute of Technology (2011-15). During 2014-15, he took an on leave and worked on the Amazon-Go project. He was a Visiting Researcher (2011-14) and a Research Staff Member (2010-11) at IBM Research T. J. Watson Center, a Senior Researcher (2009-10) at Nokia Research Center Hollywood, and a Scientist (2006-09) at Microsoft Live labs Research. He is an associate editor of TIP, TCSVT, CVIU, IEEE Multimedia, TVCJ and MVA. He also served as the Lead Guest Editor on two special issues in TPAMI and IJCV, respectively. He will be a general chair for ICCV'2027. He is a program chair of CVPR’2019 and 2022. He is a Senior Area Chair of CVPR2023 and NeurIPS2024, an area chair of CVPR’2015\&2017\&2021\&2025, ICCV’2011\&2017\&2023\&2025, ECCV'2020\&2022\&2024, ICIP’2012\&2013\&2016, ICASSP’2012\&2013, and ACM MM 2011\&2012\&2015\&2017. He is the author of more than 260 peer reviewed publications in prestigious international journals and conferences. He holds 37 US/International patents and has 15 more US patents pending. He is the recipient of the 2015 IAPR Young Biometrics Investigator Award for his contribution on Unconstrained Face Recognition from Images and Videos, and a recipient of the 2013 Google Research Faculty Award. He is an IEEE Fellow, an IAPR Fellow, and an ACM Distinguished Scientist.
\end{IEEEbiography}


%\vfill


\end{document}